%% file: r2a.tex
\documentclass[11pt,a4paper]{article}
\usepackage[hyperref]{emnlp2018}
\usepackage{times}
\usepackage{bm}
\usepackage{latexsym}
\usepackage{graphicx}
\usepackage{xcolor}
\usepackage{booktabs}
\usepackage{amsfonts}
\usepackage{amsmath}
\usepackage{multicol}
\usepackage{setspace}
\usepackage{amssymb}
\usepackage{multirow}
\usepackage{soul}
\usepackage{caption}
\usepackage{enumitem}

\usepackage{colortbl}
\usepackage{color}
\usepackage{soulpos}
\usepackage{url}
\usepackage{adjustbox}
\usepackage{subcaption}
\usepackage{ragged2e}

\usepackage{comment}

\definecolor{grey}{rgb}{0.8,0.8,0.8}
\newcommand{\hlr}[2]{\setlength{\fboxsep}{0.3pt}\colorbox{red!#2}{\rule[-.05\baselineskip]{0pt}{.7\baselineskip}{#1}}}

\newcommand{\rtoa}{\textsc{R2A}}
\newcommand{\trans}{\textsc{Trans}}
\newcommand{\ratrans}{\textsc{Ra-Trans}}
\newcommand{\ours}{\textsc{Ours}}
\newcommand{\rasvm}{\textsc{Ra-Svm}}
\newcommand{\svm}{\textsc{Svm}}
\newcommand{\racnn}{\textsc{Ra-Cnn}}
\newcommand{\oracle}{\textsc{Oracle}}

\usepackage[colorinlistoftodos]{todonotes}

\newcommand{\rmb}[1]{\bm{\mathrm{#1}}}

\aclfinalcopy 
\def\aclpaperid{1621} 

\title{Deriving Machine Attention from Human Rationales}

\author{Yujia Bao$^1$, Shiyu Chang$^{2}$, Mo Yu$^{2}$, Regina Barzilay$^1$\\
$^1$Computer Science and Artificial Intelligence Lab, MIT\\$^2$MIT-IBM Watson AI Lab, IBM Research\\
\{yujia, regina\}@csail.mit.edu, shiyu.chang@ibm.com, yum@us.ibm.com
}

\date{}

\begin{document}
\maketitle
\begin{abstract}
Attention-based models are successful when trained on large amounts of  data. In this paper, we demonstrate that even in the low-resource scenario, attention can be learned effectively. To this end, we start with discrete human-annotated rationales and map them into continuous attention. Our central hypothesis is that this mapping is general across domains, and thus can be transferred from resource-rich domains to low-resource ones.
Our model jointly learns a domain-invariant representation and induces the desired mapping between rationales and attention.
Our empirical results validate this hypothesis and show that our approach delivers significant gains over state-of-the-art baselines, yielding over 15\% average error reduction on benchmark datasets.\footnote{Our code and data are available at \url{https://github.com/YujiaBao/R2A}.}

\end{abstract}

\input{sections/introduction}
\input{sections/related}
\input{sections/model}
\input{sections/experiment}
\input{sections/results}
\input{sections/conclusion}

\section*{Acknowledgments}
We thank the MIT NLP group and the reviewers for their helpful discussion and comments.
This work is supported by MIT-IBM Watson AI Lab.
Any opinions, findings, conclusions, or recommendations expressed in this paper are those of the authors, and do not necessarily reflect the views of the funding organizations.
\newpage
\bibliographystyle{acl_natbib_nourl}
\bibliography{emnlp2018}
\input{sections/appendix}

\end{document}

%% file: sections/introduction.tex
\section{Introduction}
Attention-based models have become architectures of choice for many NLP tasks. In addition to significant performance gains, these models are attractive, as attention is often used as a proxy for human interpretable rationales. Their success, however, is conditioned on access to large amounts of training data. To make these models applicable in low-resource scenarios, we utilize this connection in the opposite direction. Specifically, we propose an approach to map human rationales to high-performing attention, and use this attention to guide models trained in low-resource scenarios.

\input{examples/intro.tex}

The notions of rationale and attention are closely related. Both of them highlight word importance for the final prediction. In the case of rationale, the importance is expressed as a hard selection, while attention provides a soft distribution over the words. Figure~\ref{fig:intro_example} illustrates this relatedness.
One obvious approach to improve low-resource performance is to directly use human rationales as a supervision for attention generation. The implicit assumption behind this method is that machine-generated attention should mimic human rationales. However, rationales on their own are not adequate substitutes for machine attention. Instead of providing a soft distribution, human rationales only provide the binary indication about relevance. Furthermore, rationales are subjectively defined and often vary across annotators. Finally, human rationales are not customized for a given model architecture. In contrast, machine attention is always derived as a part of a specific model architecture.

To further understand this connection, we empirically compare models informed by human rationales and those by high-quality attention. To obtain the latter, we derive an ``oracle'' attention using a large amount of annotations. This ``oracle'' attention is then used to guide a model that only has access to a small subset of this training data. Not only does this model outperform the oracle-free variant, but it also yields substantial gains over its counterpart trained with human rationales --- 89.98 \% vs 85.22 \% average accuracy on three aspects of hotel review (see Section~\ref{tab:res_outdomain} for details). In practice, however, this ``oracle''  attention is not available. To employ this method, we need to find a way to obtain a substitute for the ``oracle'' attention.

In this paper, we show how to achieve this goal using rationales. Specifically, we learn a mapping from human rationales to high-quality attention (R2A). We hypothesize that this mapping is generalizable across tasks and thus can be transferred from resource-rich tasks.\footnote{In this paper, we consider a more general setting where one domain contains multiple tasks. Also, we assume having one source domain. However, our proposed method is a general framework and can be easily adapted to problems with multiple source domains.} Figure~\ref{fig:intro_example} illustrates that in both tasks, attention weighs rationale words in a similar fashion: highlighting task-specific nouns and adjectives, while downplaying functional words.  To learn and apply this mapping we need access to rationales in both source and target tasks. In the target task, we assume rationales are provided by humans. In the source task(s), collecting rationales at scale is infeasible. Therefore, we use machine-generated rationales~\cite{lei2016rationalizing} as a proxy.

Our R2A model consists of three components. The first one is an attention-based model for the source task(s) that provides supervision for attention generation. The second component focuses on learning a domain-invariant representation to support transfer. The third component combines this invariant representation and rationales together to generate the attention. These three components are trained jointly to optimize the overall objective.  Once the model is trained, we apply it to the target task to generate attention from human rationales. This attention is consequently used to supervise the training of the target classifier.

We evaluate our approach on two transfer settings: aspect transfer within single domain and domain transfer across multiple domains. Our experiments demonstrate that our approach delivers significant performance improvements over the baselines.  For instance, the average error reduction over the best baseline in domain transfer is over 15\%.  In addition, both qualitative and quantitative analyses confirm that our R2A model is capable of generating high-quality attention for target tasks.

%% file: examples/intro.tex
\begin{table}[t!]
	\small
	\centering

	\begin{tabular}{p{.9\linewidth}}
		\emph{Task: Hotel location}\hspace*{0pt}\hfill label: negative\\
		\midrule
			\hlr{a}{0} \hlr{nice}{0} \hlr{and}{0} \hlr{clean}{0} \hlr{hotel}{0} \hlr{to}{0} \hlr{stay}{2} \hlr{for}{0} \hlr{business}{2} \hlr{and}{0} \hlr{leisure}{5} \hlr{.}{0} \hlr{\ul{\textbf{but}}}{0} \hlr{\ul{\textbf{the}}}{1} \hlr{\ul{\textbf{location}}}{46} \hlr{\ul{\textbf{is}}}{11} \hlr{\ul{\textbf{not}}}{70} \hlr{\ul{\textbf{good}}}{50} \hlr{\ul{\textbf{if}}}{3} \hlr{\ul{\textbf{you}}}{0} \hlr{\ul{\textbf{need}}}{1} \hlr{\ul{\textbf{public}}}{13} \hlr{\ul{\textbf{transport}}}{12} \hlr{.}{0} \hlr{it}{1} \hlr{took}{3} \hlr{too}{6} \hlr{long}{20} \hlr{for}{3} \hlr{transport}{15} \hlr{and}{6} \hlr{waiting}{12} \hlr{for}{1} \hlr{bus}{20} \hlr{.}{0} \hlr{but}{0} \hlr{the}{0} \hlr{swimming}{0} \hlr{pool}{0} \hlr{looks}{0} \hlr{good}{3} \hlr{.}{0} \\

		\arrayrulecolor{black}
		\vspace{0.1cm}
		\emph{Task: Beer aroma}\hspace*{0pt}\hfill label: positive\\
		\midrule
		\hlr{poured}{0} \hlr{a}{0} \hlr{deep}{0} \hlr{brown}{0} \hlr{color}{0} \hlr{with}{0} \hlr{little}{0} \hlr{head}{0} \hlr{that}{0} \hlr{dissipated}{6} \hlr{pretty}{7} \hlr{quickly}{28} \hlr{.}{24} \hlr{\ul{\textbf{aroma}}}{70} \hlr{\ul{\textbf{is}}}{15} \hlr{\ul{\textbf{of}}}{5} \hlr{\ul{\textbf{sweet}}}{64} \hlr{\ul{\textbf{maltiness}}}{46} \hlr{\ul{\textbf{with}}}{3} \hlr{\ul{\textbf{chocolate}}}{33} \hlr{\ul{\textbf{and}}}{0} \hlr{\ul{\textbf{caramel}}}{10} \hlr{\ul{\textbf{notes}}}{0} \hlr{.}{0} \hlr{flavor}{0} \hlr{is}{0} \hlr{also}{0} \hlr{of}{0} \hlr{chocolate}{1} \hlr{and}{0} \hlr{caramel}{0} \hlr{maltiness}{0} \hlr{.}{0} \hlr{mouthfeel}{0} \hlr{is}{0} \hlr{good}{0} \hlr{a}{0} \hlr{bit}{0} \hlr{on}{0} \hlr{the}{0} \hlr{thick}{0} \hlr{side}{0} \hlr{.}{0} \hlr{drinkability}{0} \hlr{is}{0} \hlr{ok}{0} \hlr{.}{0} \hlr{this}{0} \hlr{is}{0} \hlr{to}{0} \hlr{be}{0} \hlr{savored}{0} \hlr{not}{0} \hlr{sessioned}{0} \hlr{.}{0}\\
	\end{tabular}
	\captionof{figure}{Examples of rationales versus oracle attention. Words are highlighted according to their relative attention scores. Human rationales are shown in bold with underlines.} \label{fig:intro_example}
\end{table}

%% file: sections/related.tex
\section{Related Work}
\paragraph{Attention-based models} Attention has been shown to be effective when the model is trained on large amounts of training data~\cite{bahdanau2014neural, luong2015effective, rush2015neural, yang2016hierarchical, lin2017structured, chen2017reading, NIPS2017_7181}. In this setting, typically no additional supervision is required for learning the attention. Nevertheless, further refining attention by extra supervision has been shown to be beneficial. Examples include using word alignments to learn attention in neural machine translation~\cite{liu2016neural}, employing argument words to supervise attention in event detection~\cite{liu2017exploiting}, utilizing linguistically-motivated annotations to guide attention in constituency parsing~\cite{kamigaito2017supervised}. These supervision mechanisms are tailored to specific applications. In contrast, our approach is based on the connection between rationales and attention, and can be used for multiple applications.

\paragraph{Rationale-based models} \citet{zaidan2007using} was the first to explore the value of rationales in low-resource scenarios. They hypothesize that the model confidence should decrease when the rationale words are removed from the inputs, and validate this idea for linear models. Recent work~\cite{zhang2016rationale} explores the potential of integrating rationales with more complex neural classifiers. In their model, human rationales are directly used to guide the sentence-level attention for a CNN-based classifier. To reach good performance, their model still requires a sufficient amount of training data. Our work differs from theirs as we discern the intrinsic difference between human rationales and machine attention. Moreover, we learn a model to map human rationales into high-quality attention so as to provide a richer supervision for low-resource models. 

\paragraph{Transfer learning}
When labeled data on the target task is available, existing approaches typically transfer the knowledge by either fine-tuning an encoder trained on the source tasks(s)~\cite{conneau2017supervised, peters2018deep} or multi-task learning on all tasks with a shared encoder~\cite{collobert2011natural}. In this paper, we explore the transferability of the task-specific attention through human rationales. We believe this will further assist learning in low-resource scenarios.

Our work is also related to unsupervised domain adaptation, as the R2A model has never seen any target annotations during training. Existing methods commonly adapt the classifier by aligning the representations between the source and target domains~\cite{glorot2011domain, chen2012marginalized, zhou2016bi,ganin2016domain,zhang2017aspect}. In contrast, our model adapts the mapping from rationales to attention; thus after training, it can be applied to different target tasks.

%% file: sections/model.tex
\section{Method}
\label{sec:method}

\paragraph{Problem formulation} We assume that we have $N$ source tasks $\{\mathcal{S}_i\}_{i=1}^N$, where each of them has sufficient amounts of labeled examples. Using existing methods~\cite{lei2016rationalizing}, we can generate rationales for each source example automatically (see \hyperref[sec:a1]{Appendix 1} for details). In the target task $\mathcal{T}$, we only have a limited amount of labeled examples with large amounts of unlabeled data. For those labeled examples, we assume access to human-annotated rationales.

\begin{figure}[t!]
  \centering
  \includegraphics[width=\linewidth]{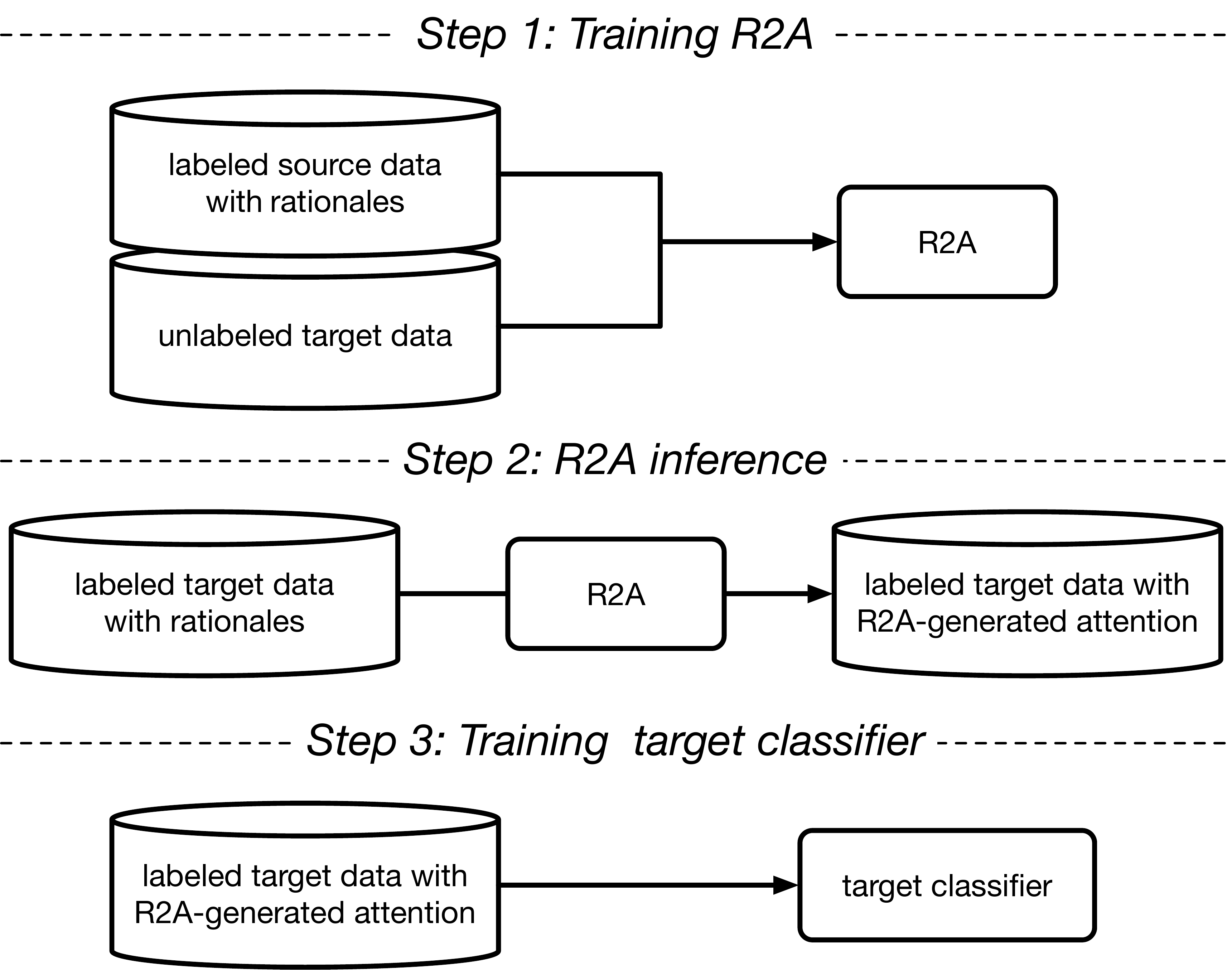}
  \caption{Overall pipeline of our approach (Section~\ref{sec:pipeline}). The R2A mapping is learned from labeled source data and unlabeled target data. Then we applied it to the target task to derive attention based on human rationales. Finally, a target classifier is trained under the supervision of both the annotated labels and the R2A-generated attention.}\label{fig:pipeline}
\end{figure}

\paragraph{Overview}
Our goal is to improve classification performance on the target task by learning a mapping from human rationales to high-quality machine attention (\rtoa). Given the scarcity of our target data, we learn this mapping on resource rich tasks where high-quality attention can be readily obtained during training. Next, the mapping between rationales and attention derived from the source tasks is exported into the target task. To enable this transfer, models have to operate over an invariant representation which we construct via an adversarial objective. Once the mapping is derived, we can translate human rationales in the target task into high-quality attention. This generated attention is then used to provide additional training signal for an attention-based classifier for the target task. The overall pipeline is shown in Figure~\ref{fig:pipeline}.

Alternatively, we can view the \rtoa\ mapping as a meta model that produces a prior over the attention distribution across different tasks.

\begin{figure*}[t!]
  \centering
  \includegraphics[width=\linewidth]{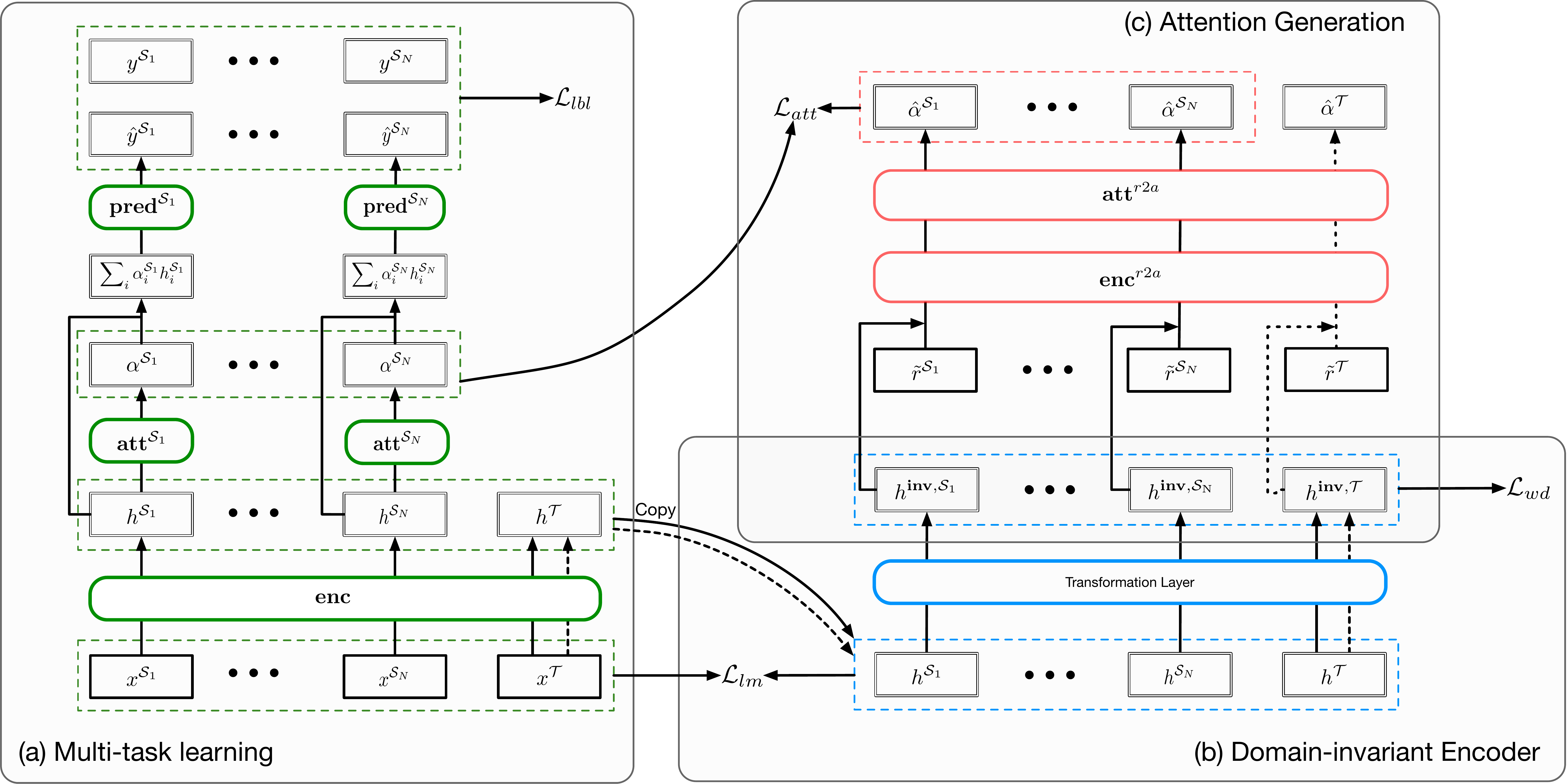}
  \caption{Architecture of the R2A model. The model is comprised of (a) a multi-task learning component, (b) a domain-invariant encoder, and (c) an attention generation component. Solid arrows denote computations for training, while dotted arrows denote computations for inference.}\label{fig:r2a}
  \vspace{-.15in}
\end{figure*}

\paragraph{Model architecture}
Figure~\ref{fig:r2a} illustrates the architecture of our \rtoa\ model, which consists of three components.
\begin{itemize}[leftmargin=*]
\item \textbf{Multi-task learning}\hspace{.1in} In order to learn the \rtoa\ mapping, we need annotation for the attention. This module generates high-quality attention as an intermediate result by minimizing the prediction error on the source tasks  (Section~\ref{ssec:meta_learn}).

\item \textbf{Domain-invariant encoder}\hspace{.1in}
This module aims to transform the contextualized representation obtained from the first module into a domain-invariant version. We achieve this goal through domain adversarial training over the source data and the unlabeled target data (Section~\ref{ssec:encoder}).

\item \textbf{Attention generation}\hspace{.1in} This module learns to predict the intermediate attention obtained from the first module based on the domain-invariant representation and the rationales (Section~\ref{ssec:r2a}).

\end{itemize}

\subsection{Multi-task learning}
\label{ssec:meta_learn}
The goal of the multi-task learning module is to learn good attention for each source task.  This learned attention will be used later to supervise the attention generation module. This module takes the input text from the source tasks and predicts the labels. To accomplish the previously stated goal, we minimize the prediction error over all labeled source data.

Let $(x^t,y^t)$ be a training instance from any source task $t\in\{\mathcal{S}_1,\ldots\mathcal{S}_N\}$. We first encode the input sequence $x^t$ into hidden states: $h^t=\rmb{enc}(x^t)$, where $\rmb{enc}$ is a bi-directional LSTM~\cite{hochreiter1997long} that is shared across all source tasks. For each position $i$, the dense vector $h_i^t$ encodes the content and context information of the word $x^t_i$. We then pass the sequence $h^t$ on to a task-specific attention module to produce attention $\alpha^t = \rmb{att}^t(h^t)$ as follows: 
\begin{equation*}
    \begin{aligned}
    \tilde{h}_i^t &= \tanh(W^t_{\rmb{att}} h_i^t+b^t_{\rmb{att}}),\\
    \alpha_i^t &= \frac{\exp(\langle \tilde{h}_i^t, q^t_{\rmb{att}} \rangle)}{\sum_j \exp (\langle\tilde{h}_j^t, q^t_{\rmb{att}} \rangle)},
    \end{aligned}
\end{equation*}
where $\langle \cdot,\,\cdot\rangle$ denotes inner product and $W^t_{\rmb{att}}$, $b^t_{\rmb{att}}$, $q^t_{\rmb{att}}$ are learnable parameters.  We predict the label of $x^t$ using the weighted sum of its contextualized representation: $\hat{y}^t=\rmb{pred}^t (\sum_i \alpha_i^t h_i^t)$, where $\rmb{pred}^t$ is a task-specific multi-layer perceptron. We train this module to minimize the loss, $\mathcal{L}_{lbl}$, between the prediction and the annotated label for all source tasks. We use cross entropy loss for classification tasks and mean square loss for regression tasks.

\subsection{Domain-invariant encoder}
\label{ssec:encoder}

Supplied with large amounts of source data and unlabeled target data, this module has two goals: 1) learning a general encoder for both source and target corpora, and 2) learning domain-invariant representation.
This module enables effective transfer---especially in the presence of significant variance between the source and target domains.
We achieve the first goal by optimizing a language modeling objective and the second goal by minimizing the Wasserstein distance between the source and target distribution.

Let $x$ be an input sequence, and $h\triangleq[\overrightarrow{h};\overleftarrow{h}]$ be its corresponding contextualized representation obtained from $\rmb{enc}$.
Here, $\overrightarrow{h}$ and $\overleftarrow{h}$ denote the output sequence of the forward and backward LSTM, respectively.
In order to support transfer, $\rmb{enc}$ should be general enough to effectively represent both source and target corpora.  For this reason, we ground the encoder by a language modeling component~\cite{bengio2003neural, mikolov2011extensions}.
Specifically, we employ two Softmax classifiers to predict the word $x_i$ based on $\overrightarrow{h}_{i-1}$ and $\overleftarrow{h}_{i+1}$ respectively.
We minimize the cross-entropy loss $\mathcal{L}_{lm}$ over all source data and unlabeled target data.

The representation $h$ is domain-specific as it is trained to encode useful features for language modeling and the source tasks. To obtain an invariant representation, we employ a transformation layer and propose to align the transformed representation so that it is not distinguishable whether it comes from the source or the target.
Specifically, we transform the representation $h_i$ at each position $i$ linearly and obtain
\begin{equation*}
    h^{\rmb{inv}}_i = W_{\rmb{inv}}h_i + b_{\rmb{inv}},
\end{equation*}
where $W_{\rmb{inv}}$ and $b_{\rmb{inv}}$ are learnable parameters.
We minimize the Wasserstein distance~\cite{arjovsky2017wasserstein} between the distribution of $h^{\rmb{inv}}$ from the source and the one from the target, denoted as $\mathbb{P}_{\mathcal{S}}$ and $\mathbb{P}_{\mathcal{T}}$, respectively. 
Since $h^{\rmb{inv}}$ is a sequence of variable length, $L$, we summarize it by its first and last element via concatenation, $[h^{\rmb{inv}}_1; h^{\rmb{inv}}_{L}]$. 
The training objective is defined as: 
\begin{equation*}
    \begin{aligned}
    \mathcal{L}_{wd} = \sup_{\|f\|_{L}\leq K}
    &\mathbb{E}_{h^{\rmb{inv}} \sim
    \mathbb{P}_{\mathcal{S}}}\left[f([h^{\rmb{inv}}_1; h^{\rmb{inv}}_{L}])\right]\\
    \quad &-\mathbb{E}_{h^{\rmb{inv}} \sim
    \mathbb{P}_{\mathcal{T}}}\left[f([h^{\rmb{inv}}_1; h^{\rmb{inv}}_{L}])\right],
    \end{aligned}
\end{equation*}
where the supremum is over all $K$-Lipschitz scalar functions $f$. Following \citet{gulrajani2017improved}, we approximate $f$ by a multi-layer perceptron, and use gradient penalty to fulfill the Lipschitz constraint.

\subsection{Attention generation}
\label{ssec:r2a}
The goal of this module is to generate high-quality attention for each task.  This module combines the domain-invariant representation together with task-specific rationales as its input and predicts task-specific attention scores.  We minimize the distance between the predicted attention and the intermediate attention obtained from the multi-task learning module.

For any source task $t\in\{\mathcal{S}_i\}_{i=1}^N$, we denote $r^t$ as the task-specific rationales corresponding to the input text $x^t$, and denote $h^{\rmb{inv},t}$ as the domain-invariant representation of $x^t$.  For each position $i$, we first concatenate $r^t_i$ with the frequency of $x_i^t$ occurring as a rationale from all training examples of this task.  We denote this augmented sequence as $\tilde{r}^t$.  This frequency term provides the unigram likelihood of each word being a rationale for the task.  Then we employ a sequence encoder $\rmb{enc}^{r2a}$ and an attention module $\rmb{att}^{r2a}$ to predict the attention scores:
\begin{equation*}
    \begin{aligned}
        u^t &= \rmb{enc}^{r2a}([h^{\rmb{inv},t}; \tilde{r}^t]),\\
        \tilde{u}_i^t &= \tanh(W_{\rmb{att}}^{r2a} u_i^t +
        b_{\rmb{att}}^{r2a}), \\
        \hat{\alpha}_i^t &= \frac{\exp(\langle \tilde{u}_i^t, q_{\rmb{att}}^{r2a} \rangle)}
        {\sum_j\exp(\langle \tilde{u}_j^t, q_{\rmb{att}}^{r2a} \rangle)},
    \end{aligned}
\end{equation*}
where $W_{\rmb{att}}^{r2a}$, $b_{\rmb{att}}^{r2a}$ and $q_{\rmb{att}}^{r2a}$ are learnable parameters, and both $\rmb{enc}^{r2a}$ and $\rmb{att}^{r2a}$ are shareable across all tasks.  We minimize the distance between $\hat{\alpha}^t$ and the  $\alpha^t$ obtained from the first multi-task learning module over all source data:
\begin{equation*}
    \mathcal{L}_{att} =
    \sum_{(\alpha^t,\hat{\alpha}^t), t\in \{\mathcal{S}_i\}_{i=1}^N} \mathrm{d}(\alpha^t,
    \hat{\alpha}^t), \label{eq:reconstruct_loss}
\end{equation*}
where  $\mathrm{d}(\cdot,\cdot)$ can be any distance metric. In this paper, we consider a soft-margin cosine distance:
\[
\mathrm{d}(a,b)\triangleq\max(0,1-\cos(a,b)-0.1),
\]
where $\cos(\cdot,\cdot)$ denotes the cosine similarity.

\subsection{Pipeline}
\label{sec:pipeline}
\paragraph{Training R2A} We train the three aforementioned modules jointly using both the source data and the unlabeled target data. The overall objective is given by:
\begin{equation}
\mathcal{L}= \mathcal{L}_{lbl} + 
\lambda_{att}\mathcal{L}_{att} +
\lambda_{lm}\mathcal{L}_{lm} +
\lambda_{wd}\mathcal{L}_{wd}.
\end{equation}
The $\lambda$s are hyper-parameters that control the importance of each training objective and $\mathcal{L}$s represent the corresponding loss functions.

\paragraph{R2A inference} 
Once the \rtoa\ model is trained, we can generate attention for each labeled target example based on its human-annotated rationales.

\paragraph{Training target classifier} 
When testing the performance on the target task, of course, we are neither provided with labels nor rationales. In order to make predictions for the target task, we train a new classifier under the supervision of both the labels and the R2A-generated attention. Specifically, this target classifier shares the same architecture as the source one in the multi-task learning module. We minimize the prediction loss, $\mathcal{L}_{lbl}^{\mathcal{T}}$, on the labeled target data together with the cosine distance, $\mathcal{L}_{att}^{\mathcal{T}}$, between the R2A-generated attention and the attention generated by this target classifier.  
The joint objective for this target classifier is defined as
\begin{equation}\label{eq:transfer_loss}
\mathcal{L} = \mathcal{L}_{lbl}^\mathcal{T} + \lambda_{att}^{\mathcal{T}}\mathcal{L}_{att}^{\mathcal{T}},
\end{equation}
where $\lambda_{att}^{\mathcal{T}}$ controls the importance of $\mathcal{L}_{att}^{\mathcal{T}}$.  For better transfer, we initialize the encoder in the target classifier as $\rmb{enc}$ from the trained R2A model. 

%% file: sections/experiment.tex
\section{Experimental Setup}
\subsection{Datasets}

We evaluate our approach on two transfer settings:  transfer among aspects within the same domain and transfer among different domains.

\paragraph{Aspect transfer}
We first consider the transfer problem between multiple aspects of one domain: beer review.  We use a subset of the BeerAdvocate\footnote{\url{https://www.beeradvocate.com}} review dataset~\cite{mcauley2012learning} introduced by~\citet{lei2016rationalizing}. This dataset contains reviews with ratings (in the scale of $[0,1]$) from three aspects of the beer: look, aroma and palate. We treat any two aspects as the source and the other one as the target.  
We consider a classification setting for each target task.  Specifically, reviews with ratings $\leq 0.4$ are labeled as negative and those with $\geq 0.6$ are labeled as positive.  We form our dataset by randomly selecting an equal number of positive and negative examples.  Then we randomly select 200 examples and ask human annotators to provide rationales (see \hyperref[sec:a2]{Appendix 2} for details).  These 200 examples are treated as our labeled training data for the target aspect. Unlabeled target data is not required since both source and target tasks are from the same domain. Table~\ref{tab:data_beer} summarizes the statistics of the beer review dataset.

\begin{table}[t!]
	\small
    \centering
    \begin{tabular}{lrrrrr}
        \toprule
        \shortstack[l]{Beer\\Aspects}   & \shortstack[c]{Source\\Train} & \shortstack[c]{Source\\Dev} & \shortstack[c]{Target\\Train$^\ddagger$} & \shortstack[c]{Target\\Dev} & \shortstack[c]{Target\\Test}\\
		\midrule
		Look  	  & 43,351     & 10,170 &   200 & 200 & 4,014\\
		Aroma     & 39,825     & 8,772  &   200 & 200 & 4,212\\
		Palate    & 30,041     & 7,152  &   200 & 200 & 3,804\\
        \bottomrule
    \end{tabular}
    \caption{Statistics of the beer review dataset. $^\ddagger$ denotes data with human-annotated rationales.}\label{tab:data_beer}
\end{table}

\paragraph{Domain transfer}
Our second experiment focuses on domain transfer from beer reviews to different aspects of hotel reviews.  We use the TripAdvisor\footnote{\url{https://www.tripadvisor.com}} hotel review dataset~\cite{wang2010latent}, with the following three aspects as our transfer target: location, cleanliness, and service.  For each aspect, reviews with ratings $>3$ are labeled as positive and those with $<3$ are labeled as negative. Similarly, we collect human rationales for 200 examples and treat them as our training data (see \hyperref[sec:a2]{Appendix 2} for details). Table~\ref{tab:data_hotel} summarizes the statistics of the hotel review dataset. In this experiment, data from all three aspects of the beer reviews are treated as the source tasks.

\begin{table}[t!]
	\small
    \centering
    \begin{tabular}{lccrc}
        \toprule
        \shortstack[l]{Hotel\\Aspects}   & \shortstack[c]{Target\\Train$^\ddagger$} & \shortstack[c]{Target\\Dev} & \shortstack[c]{Target\\Test} & \shortstack[c]{Target\\Unlabeled}\\
		\midrule
        Location    & 200 & 200 & 1,808 & 14,472\\
        Cleanliness & 200 & 200 & 12,684 & 14,472\\
        Service     & 200 & 200 & 18,762 & 14,472\\
        \bottomrule
    \end{tabular}
    \caption{Statistics of the hotel review dataset. $^\ddagger$ denotes data with human-annotated rationales.}\label{tab:data_hotel}
        	\vspace{-.1in}
\end{table}
\subsection{Baselines}
We compare our approach (\ours) with four types of baselines:

\paragraph{Basic classifier}
We train a linear \svm\ using bag-of-ngrams representation on the labeled target data. We combine uni-gram, bi-grams, and tri-grams as features and use tf-idf weighting.

\paragraph{Rationale augmented classifiers}
We evaluate two previous methods that incorporate human rationales during training: 1) rationale augmented SVM (\rasvm)~\cite{zaidan2007using}, an SVM-based model that utilizes human rationales to regularize the decision boundary of the classifier; 2) rationale augmented CNN (\racnn)~\cite{zhang2016rationale}. \racnn\ first trains a CNN-based sentence classifier to estimate the probability that a given sentence contains rationale words.  Then \racnn\ scales the contribution of each sentence to the overall representation in proportion to these estimates. The final prediction is made from this overall representation.

\paragraph{Transfer methods}
We compare against two variants of our method: 1) \trans, an attention-based classifier that does not use human rationales from the target task; 2) \ratrans, an attention-based classifier that directly uses human rationales to supervise the attention. Specifically, \trans\ only optimizes the cross-entropy loss $\mathcal{L}_{lbl}^{\mathcal{T}}$ in the objective (Eq.~\eqref{eq:transfer_loss}). For \ratrans, the term $\mathcal{L}_{att}^{\mathcal{T}}$ in the objective Eq.~\eqref{eq:transfer_loss} is replaced by the cosine distance between human rationales and the attention generated by itself. Note that both models still have the ability to transfer, as their encoders are both initialized from $\rmb{enc}$, which has been trained on source data and unlabeled target data.

\paragraph{Oracle}
We also report the performance of an \oracle\ which shares the same architecture as ours but is supervised by the oracle attention. The oracle attention is derived from a held-out dataset with large-scale annotations for the target task (see \hyperref[sec:a3]{Appendix 3} for details). This helps us analyze the contribution of our R2A approach in isolation of the inherent limitations of the target tasks.

\begin{table*}[t!]
	\small
    \centering
    \begin{tabular}{llccccccc}
        \toprule
        Source & Target & \svm & \rasvm$^\ddagger$ & \racnn$^\ddagger$ & \trans$^\dagger$ & \ratrans $^{\ddagger\dagger}$ & \ours $^{\ddagger\dagger}$ & \oracle$^{\dagger}$\\
        \midrule
        Beer aroma+palate& Beer look & 74.41	&74.83	&74.94	&72.75	&76.41&	\textbf{79.53}&	80.29 \\
        \midrule
        Beer look+palate & Beer aroma & 68.57	&69.23	&67.55	&69.92	&76.45	&\textbf{77.94}	&78.11\\
        \midrule
        Beer look+aroma & Beer palate &63.88	&67.82	&65.72	&74.66	&73.40	&\textbf{75.24}	&75.50\\
        \bottomrule
    \end{tabular}
    \caption{Accuracy of transferring between aspects. Models with $\dagger$ use labeled data from source aspects. Models with $\ddagger$ use human rationales on the target aspect.}\label{tab:res_indomain}
\end{table*}

\begin{table*}[t!]
	\small
		\vspace{-.1in}
    \centering
    \begin{tabular}{llccccccc}
        \toprule
        Source & Target & \svm & \rasvm$^\ddagger$ & \racnn$^\ddagger$ & \trans$^\dagger$ & \ratrans$^{\ddagger\dagger}$ & \ours$^{\ddagger\dagger}$ & \oracle$^{\dagger}$\\
        \midrule
        \multirow{4}{*}{\shortstack[l]{Beer look\\\qquad+\\Beer aroma\\\qquad+\\Beer palate}} & Hotel location & 78.65	& 79.09	& 79.28	& 80.42	& 82.10	& \textbf{84.52} & 85.43\\
        \cmidrule(lr){2-9}
        & Hotel cleanliness & 86.44&	86.68	&89.01	&86.95	&87.15	&\textbf{90.66}	&92.09 \\
        \cmidrule(lr){2-9}
        & Hotel service & 85.34	&86.61&	87.91	&87.37	&86.40	&\textbf{89.93}	&92.42\\
        \bottomrule
    \end{tabular}
    \caption{Accuracy of transferring between domains. Models with $\dagger$ use labeled data from source domains and unlabeled data from the target domain. Models with $\ddagger$ use human rationales on the target task.}\label{tab:res_outdomain}
    	\vspace{-.1in}
\end{table*}

\subsection{Implementation details}
We use pre-trained fastText embeddings~\cite{bojanowski2017enriching}, a 200-dimension bi-directional LSTM~\cite{hochreiter1997long} for the language encoder, and a 50-dimension bi-directional LSTM for the R2A encoder. Dropout~\cite{srivastava2014dropout} is applied with drop rate $0.1$ on the word embeddings and the last hidden layers of the classifiers. All parameters are optimized using Adam~\cite{kingma2014adam}. We set the initial learning rate to 0.001 and divide it by 10 once the performance on the development set plateaus. For \ratrans, \ours\ and \oracle, we tuned $\lambda_{att}^{\mathcal{T}}$ from $\{10^{2},10^1,10^0,10^{-1},10^{-2}\}$. For domain transfer, we set $\lambda_{lm}=0.1$, $\lambda_{wd}=0.01$ and $\lambda_{att}=0.01$. For aspect transfer, we adapt the same hyper-parameters, but set $\lambda_{wd}=0$ as both source tasks and the target task come from the same domain. To encourage the R2A-generated attention to be consistent with the provided rationales in aspect transfer, we augment the overall training objective of R2A in Eq.~\eqref{eq:reconstruct_loss} by a consistency regularization, which is computed from the cosine distance between the R2A-generated attention and the provided rationales. 

In addition, computing $\mathcal{L}_{lm}$ is both time and memory inefficient because the complexity is linear to the size of the vocabulary, which can be very large.  To expedite the training, we adopt a technique proposed by~\citet{mikolov2011extensions}, which randomly splits the entire vocabulary into a pre-defined number of bins and minimizes the loss of the bin prediction instead of the exact token prediction.  We set the bin size as 100 for our experiment.

%% file: sections/results.tex
\section{Results}

\paragraph{Aspect transfer}
\label{sec:aspect}
Table~\ref{tab:res_indomain} summarizes the results of aspect transfer on the beer review dataset.  Our model (\ours) obtains substantial gains in accuracy over the baselines across all three target aspects. It closely matches the performance of \oracle\ with only  0.40\% absolute difference.

Specifically, all rationale-augmented methods (\rasvm, \ratrans\, and \ours) outperform their rationale-free counterparts on average. This confirms the value of human rationales in the low-resource settings. We observe that the transfer baseline that directly uses rationale as augmented supervision (\ratrans) underperforms \oracle\ by a large margin. This validates our hypothesis that human rationales and attention are different.

\paragraph{Domain transfer}
Table~\ref{tab:res_outdomain} presents the results of domain transfer using 200 training examples. We use the three aspects of the beer review data together as our source tasks while use the three aspects of hotel review data as the target. Our model (\ours) shows marked performance improvement. The error reduction over the best baseline is 15.08\% on average.

\begin{figure*}[t!]
\centering
\begin{subfigure}[b]{0.32\linewidth}
\includegraphics[width=\linewidth]{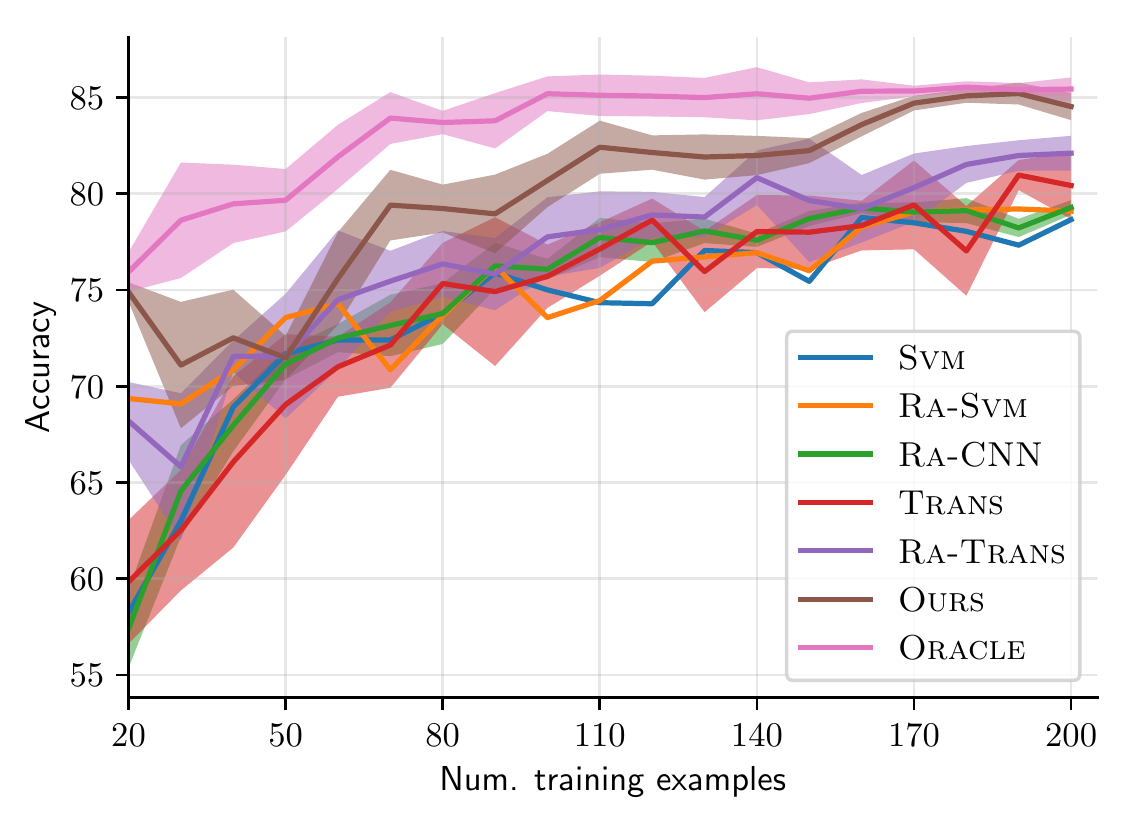}
\end{subfigure}
\begin{subfigure}[b]{0.32\linewidth}
\includegraphics[width=\linewidth]{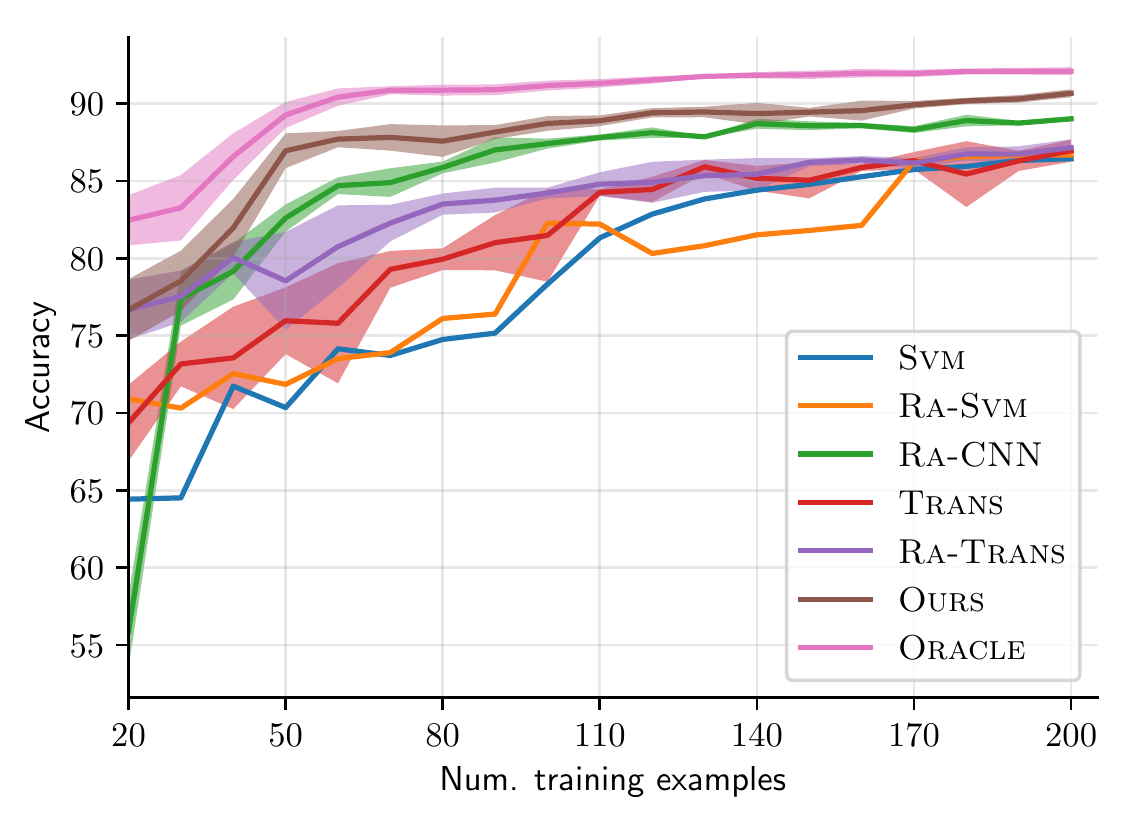}
\end{subfigure}
\begin{subfigure}[b]{0.32\linewidth}
\includegraphics[width=\linewidth]{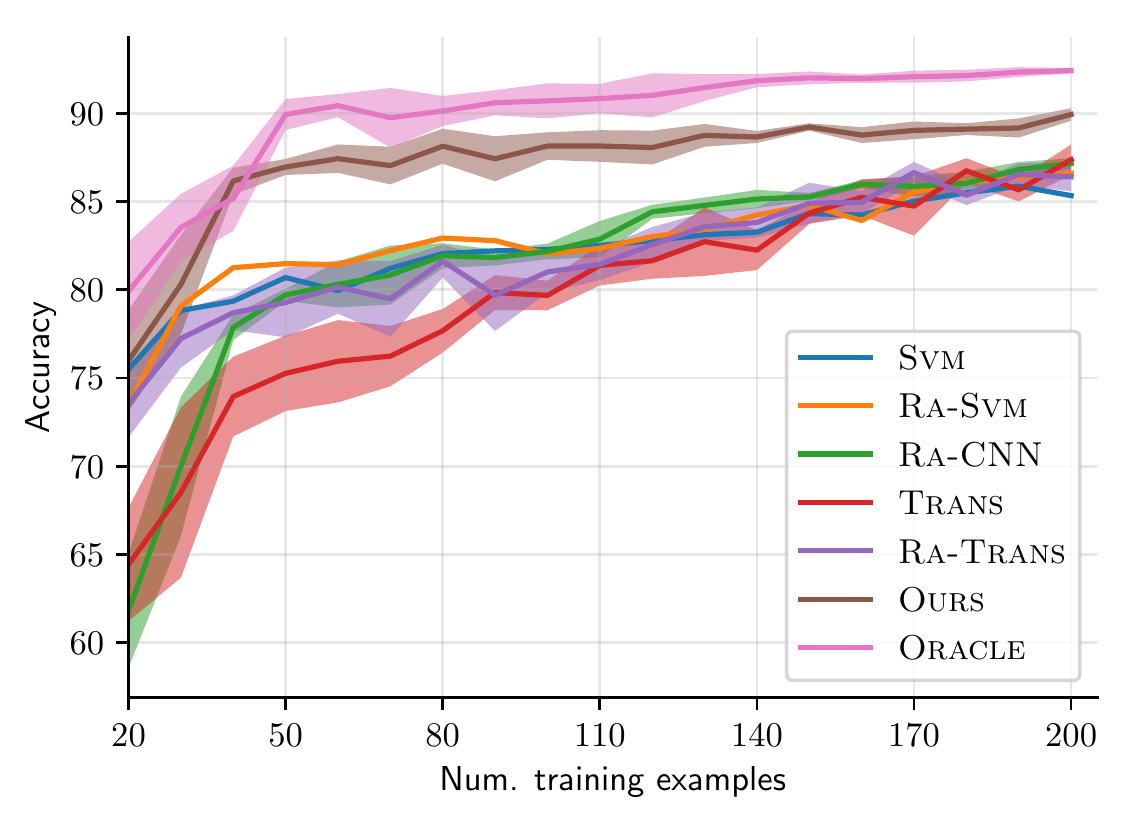}
\end{subfigure}
\caption{Learning curve of transferring from beer review to three aspects of hotel review: location (left), cleanliness (center) and service (right). For neural methods, we ran five different random seeds and plot their mean with their standard deviation.}\label{fig:learning_curve1}
\vspace{-.15in}
\end{figure*}

We compare the learning curve in Figure~\ref{fig:learning_curve1}. We observe that the performance of our model steadily improves as more annotations are provided, and the improvement over other baselines is significant and consistent.

\begin{table}[t]
	\small
    \centering
    \begin{tabular}{lccc}
        \toprule
        Model &  \shortstack[c]{Hotel\\location} & \shortstack[c]{Hotel\\cleanliness} & \shortstack[c]{Hotel\\service} \\
        \midrule
        \ours     & \textbf{84.52} & \textbf{90.66} & \textbf{89.93}\\
        \qquad w/o $\mathcal{L}_{wd}$ & 82.36 & 89.79 & 89.61 \\
        \qquad w/o $\mathcal{L}_{lm}$ & 82.47 & 90.05 & 89.75\\
        \bottomrule
    \end{tabular}
    \caption{Ablation study on domain transfer from beer to hotel.}\label{tab:ablation}
    \vspace{-.2in}
\end{table}

\begin{figure}[ht]
  \centering
  \begin{subfigure}[b]{0.49\linewidth}
    \includegraphics[width=\linewidth]{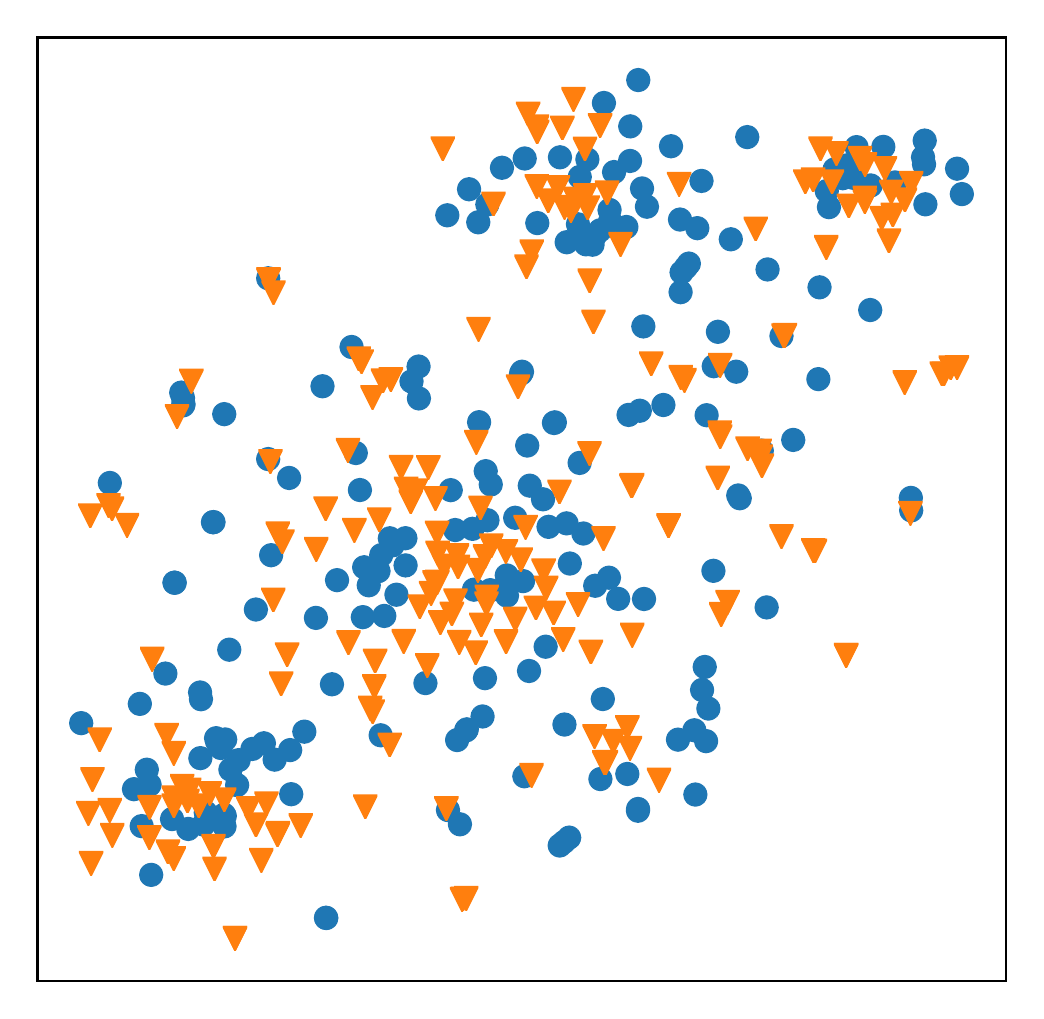}
    \caption{\ours}
  \end{subfigure}
  \begin{subfigure}[b]{0.49\linewidth}
    \includegraphics[width=\linewidth]{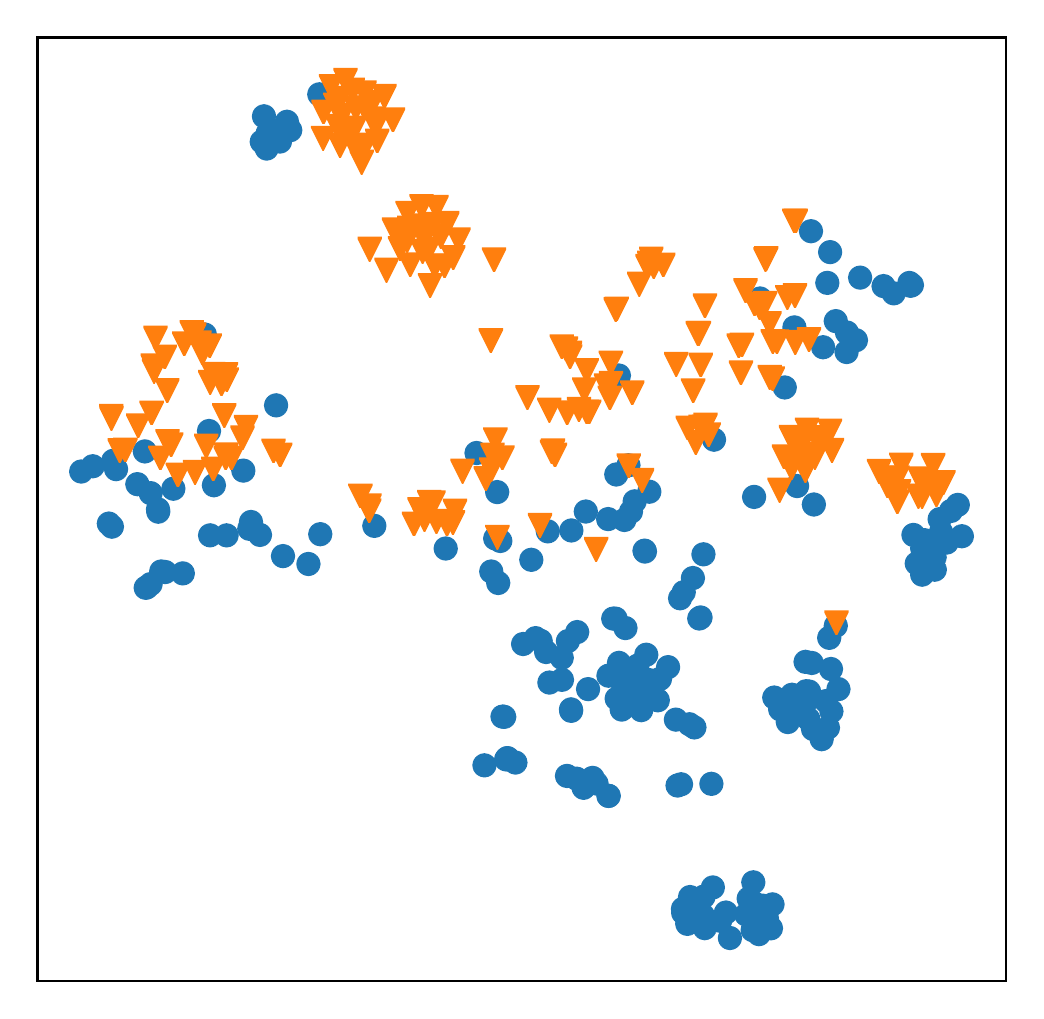}
    \caption{\ours\ w/o $\mathcal{L}_{wd}$}
  \end{subfigure}

  \caption{t-SNE visualization of the learned hidden representation\protect\footnotemark\  for beer review (blue circle) and hotel review (orange triangle).}
  \label{fig:wd}
\end{figure}

\paragraph{Ablation study}
Table~\ref{tab:ablation} presents the results of an ablation study of our model in the setting of domain transfer. As this table indicates, both the language modeling objective and the Wasserstein distance contribute similarly to the task, with the Wasserstein distance having a bigger impact.

\paragraph{Visualization of representation}

Figure~\ref{fig:wd} visualizes the hidden representation of 200 beer reviews and 200 hotel reviews using t-SNE~\cite{maaten2008visualizing}. We observe that our model successfully aligns the source and the target feature distribution. This indicates the effectiveness of optimizing the Wasserstein distance objective.

\footnotetext{Since the hidden representation is a sequence of variable length, we applied t-SNE on the concatenation of the first and last element: $[h_1^{\rmb{inv}}; h_L^{\rmb{inv}}]$.}

\begin{table}[t!]
	\small
    \centering
    \begin{tabular}{lcc}
        \toprule
        \shortstack[c]{Target} &  \shortstack[c]{Human\\rationales} & \shortstack[c]{R2A-generated\\attention} \\
        \midrule
        Location & 0.5185 & \textbf{0.2371}\\
        Cleanliness & 0.5948 & \textbf{0.3141}\\
        Service  & 0.5833 & \textbf{0.2871}\\
        \bottomrule
    \end{tabular}
    \caption{Avg. cosine distance to the oracle attention over the target training set. The R2A is trained on beer reviews with unlabeled hotel reviews.}\label{tab:cosine}
    \vspace{-.1in}
\end{table}

\paragraph{Analysis of R2A-generated attention}
In order to validate that the trained R2A model is able to generate task-specific attention from human rationales, we perform both qualitative and quantitative analysis on the R2A-generated attention in the setting of domain transfer. It is worth pointing out that our \rtoa\ model has never seen any labeled hotel reviews during training.

Table~\ref{tab:cosine} presents the average cosine distance between the R2A-generated attention and the oracle attention over the target training set.  Compared with human rationales, the R2A-generated attention is much closer to the oracle attention.  This explains the large performance boost of our method.

Figure~\ref{fig:hotel_res} visualize the R2A-generated attention on the same hotel review with human rationales corresponding to three different aspects. We observe that the trained R2A model is able to produce task-specific attention scores corresponding to the provided human rationale. For example, given the rationale sentence ``not the cleanest rooms but bed was clean and so was bathroom'', R2A recognizes that not every token is equally important, and the attention should focus more on ``clean'', ``cleanest'', ``rooms'' and ``bathroom''.

\input{examples/res_hotel.tex}
\begin{figure*}[t!]
  \centering
  \begin{subfigure}[b]{0.325\linewidth}
    \includegraphics[width=\linewidth]{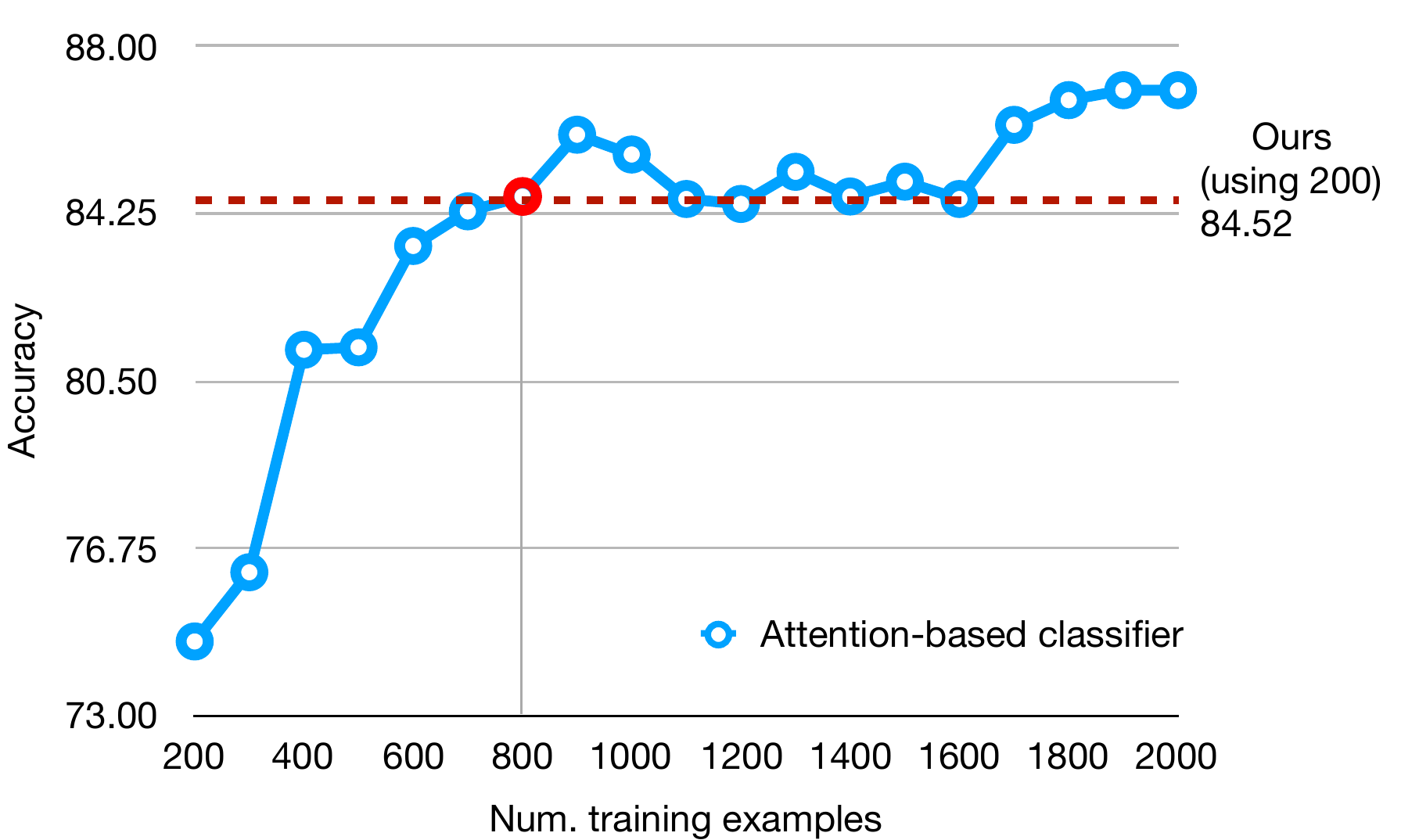}
  \end{subfigure}
  \begin{subfigure}[b]{0.325\linewidth}
    \includegraphics[width=\linewidth]{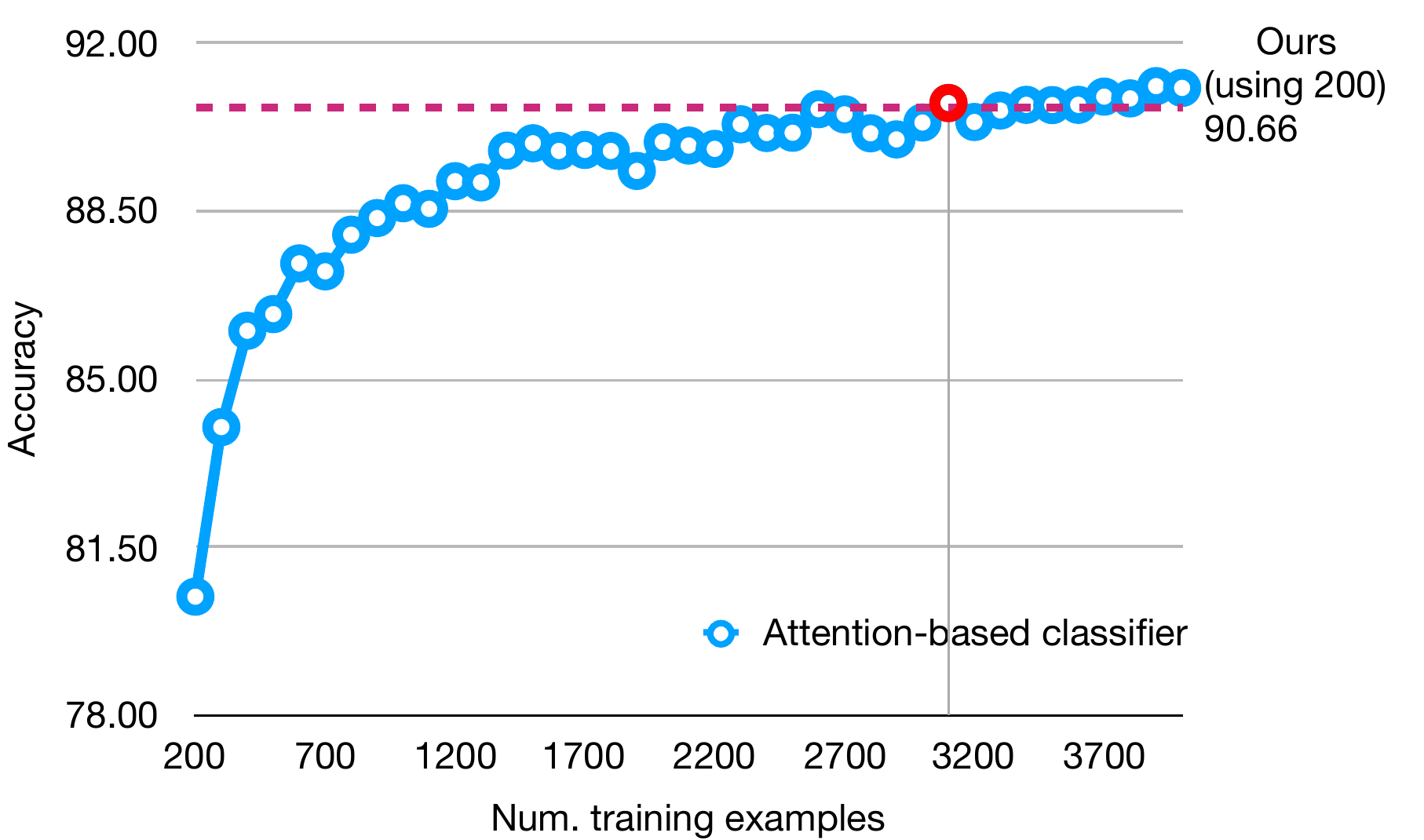}
  \end{subfigure}
    \begin{subfigure}[b]{0.325\linewidth}
    \includegraphics[width=\linewidth]{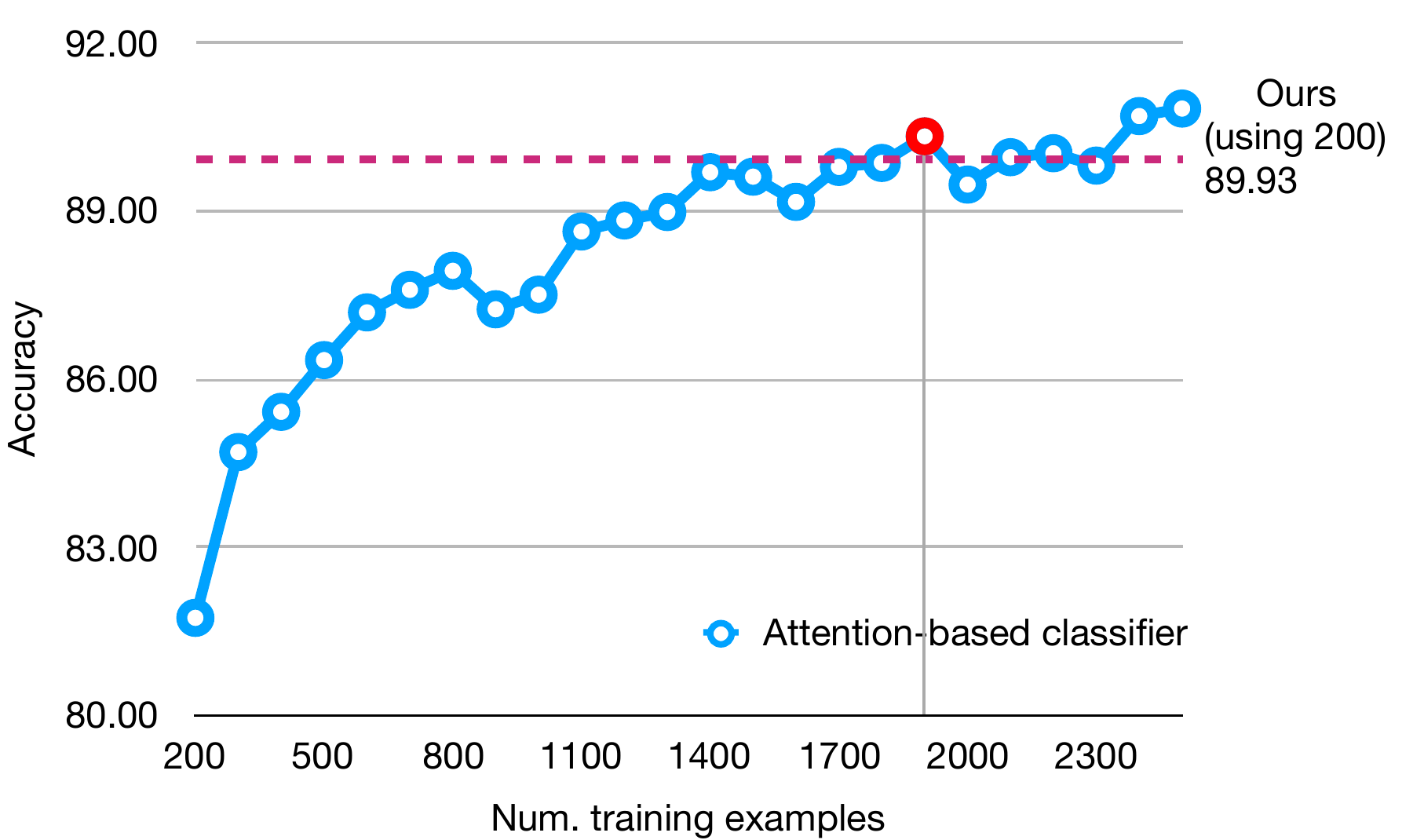}
  \end{subfigure}
  \caption{Learning curve of an attention-based classifier on three tasks: hotel location (left), hotel cleanliness (center), hotel service (right). The performance of our approach trained on 200 examples with human rationales is shown as a reference.}\label{fig:learning_curve2}
  \vspace{-.1in}
\end{figure*}

\paragraph{Annotating rationales versus annotating more labeled data}
Providing rationales for the training data roughly doubles the annotation cost~\cite{zaidan2007using}.
Given the same annotation budget, a natural question is: shall we collect a few labeled examples with rationales or annotate more labeled examples? To answer this question, we vary the number of training examples in the target task. Figure~\ref{fig:learning_curve2} shows the corresponding learning curve of a classifier that is trained without rationales. The reference line represents the accuracy of our approach trained on 200 examples with rationales. We notice that in order to reach the same level of performance, the rationale-free classifier requires 800, 3100, and 1900 labeled examples on the three target tasks respectively. 

%% file: examples/res_hotel.tex
\begin{table*}[t]
	\small
    \hspace{-0.31cm}
	\begin{tabular}{p{\linewidth}}
        \emph{Task: Hotel location}\hspace*{0pt}\hfill Oracle attention\\
		\arrayrulecolor{grey}  
		\midrule
        \hlr{you}{0} \hlr{get}{1} \hlr{what}{1} \hlr{you}{0} \hlr{pay}{1} \hlr{for}{0} \hlr{.}{0} \hlr{not}{3} \hlr{the}{1} \hlr{cleanest}{8} \hlr{rooms}{0} \hlr{but}{0} \hlr{bed}{0} \hlr{was}{0} \hlr{clean}{3} \hlr{and}{0} \hlr{so}{0} \hlr{was}{0} \hlr{bathroom}{4} \hlr{.}{0} \hlr{bring}{0} \hlr{your}{0} \hlr{own}{0} \hlr{towels}{0} \hlr{though}{0} \hlr{as}{0} \hlr{very}{0} \hlr{thin}{11} \hlr{.}{1} \hlr{service}{0} \hlr{was}{0} \hlr{excellent}{44} \hlr{,}{1} \hlr{let}{0} \hlr{us}{0} \hlr{book}{0} \hlr{in}{0} \hlr{at}{0} \hlr{8:30am}{2} \hlr{!}{0} \hlr{\ul{\textbf{for}}}{0} \hlr{\ul{\textbf{location}}}{70} \hlr{\ul{\textbf{and}}}{2} \hlr{\ul{\textbf{price}}}{1} \hlr{\ul{\textbf{,}}}{0} \hlr{\ul{\textbf{this}}}{0} \hlr{\ul{\textbf{ca}}}{0} \hlr{\ul{\textbf{n't}}}{0} \hlr{\ul{\textbf{be}}}{0} \hlr{\ul{\textbf{beaten}}}{21} \hlr{,}{0} \hlr{but}{0} \hlr{it}{0} \hlr{is}{0} \hlr{cheap}{36} \hlr{for}{3} \hlr{a}{1} \hlr{reason}{0} \hlr{.}{0} \hlr{if}{0} \hlr{you}{0} \hlr{come}{0} \hlr{expecting}{1} \hlr{the}{0} \hlr{hilton}{0} \hlr{,}{0} \hlr{then}{0} \hlr{book}{0} \hlr{the}{1} \hlr{hilton}{1} \hlr{!}{1} \hlr{for}{0} \hlr{uk}{0} \hlr{travellers}{2} \hlr{,}{0} \hlr{think}{0} \hlr{of}{1} \hlr{a}{0} \hlr{blackpool}{1} \hlr{b\&b.}{1} \\
		\vspace{0.2mm}
		\emph{Task: Hotel location}\hspace*{0pt}\hfill R2A-generated attention\\
		\arrayrulecolor{grey}  
		\midrule
        \hlr{you}{2} \hlr{get}{5} \hlr{what}{6} \hlr{you}{2} \hlr{pay}{4} \hlr{for}{2} \hlr{.}{1} \hlr{not}{1} \hlr{the}{1} \hlr{cleanest}{7} \hlr{rooms}{6} \hlr{but}{1} \hlr{bed}{2} \hlr{was}{1} \hlr{clean}{7} \hlr{and}{1} \hlr{so}{1} \hlr{was}{1} \hlr{bathroom}{4} \hlr{.}{1} \hlr{bring}{1} \hlr{your}{1} \hlr{own}{1} \hlr{towels}{2} \hlr{though}{1} \hlr{as}{1} \hlr{very}{1} \hlr{thin}{6} \hlr{.}{1} \hlr{service}{1} \hlr{was}{0} \hlr{excellent}{19} \hlr{,}{3} \hlr{let}{1} \hlr{us}{2} \hlr{book}{1} \hlr{in}{1} \hlr{at}{1} \hlr{8:30am}{2} \hlr{!}{2} \hlr{\ul{\textbf{for}}}{3} \hlr{\ul{\textbf{location}}}{70} \hlr{\ul{\textbf{and}}}{7} \hlr{\ul{\textbf{price}}}{22} \hlr{\ul{\textbf{,}}}{5} \hlr{\ul{\textbf{this}}}{4} \hlr{\ul{\textbf{ca}}}{6} \hlr{\ul{\textbf{n't}}}{4} \hlr{\ul{\textbf{be}}}{5} \hlr{\ul{\textbf{beaten}}}{60} \hlr{,}{7} \hlr{but}{5} \hlr{it}{2} \hlr{is}{3} \hlr{cheap}{28} \hlr{for}{5} \hlr{a}{2} \hlr{reason}{4} \hlr{.}{2} \hlr{if}{2} \hlr{you}{1} \hlr{come}{2} \hlr{expecting}{1} \hlr{the}{1} \hlr{hilton}{3} \hlr{,}{1} \hlr{then}{1} \hlr{book}{1} \hlr{the}{1} \hlr{hilton}{4} \hlr{!}{2} \hlr{for}{3} \hlr{uk}{4} \hlr{travellers}{3} \hlr{,}{1} \hlr{think}{1} \hlr{of}{1} \hlr{a}{1} \hlr{blackpool}{1} \hlr{b\&b.}{1} 
		\\\vspace{0.2mm}
		\emph{Task: Hotel cleanliness}\hspace*{0pt}\hfill Oracle attention\\
		\arrayrulecolor{grey}  
		\midrule
		\hlr{you}{0} \hlr{get}{0} \hlr{what}{0} \hlr{you}{1} \hlr{pay}{9} \hlr{for}{0} \hlr{.}{0} \hlr{\ul{\textbf{not}}}{0} \hlr{\ul{\textbf{the}}}{0} \hlr{\ul{\textbf{cleanest}}}{70} \hlr{\ul{\textbf{rooms}}}{5} \hlr{\ul{\textbf{but}}}{0} \hlr{\ul{\textbf{bed}}}{1} \hlr{\ul{\textbf{was}}}{1} \hlr{\ul{\textbf{clean}}}{66} \hlr{\ul{\textbf{and}}}{0} \hlr{\ul{\textbf{so}}}{1} \hlr{\ul{\textbf{was}}}{2} \hlr{\ul{\textbf{bathroom}}}{15} \hlr{.}{1} \hlr{bring}{2} \hlr{your}{1} \hlr{own}{1} \hlr{towels}{12} \hlr{though}{0} \hlr{as}{0} \hlr{very}{0} \hlr{thin}{3} \hlr{.}{0} \hlr{service}{2} \hlr{was}{0} \hlr{excellent}{16} \hlr{,}{0} \hlr{let}{1} \hlr{us}{0} \hlr{book}{3} \hlr{in}{0} \hlr{at}{0} \hlr{8:30am}{1} \hlr{!}{0} \hlr{for}{0} \hlr{location}{0} \hlr{and}{0} \hlr{price}{1} \hlr{,}{0} \hlr{this}{0} \hlr{ca}{0} \hlr{n't}{0} \hlr{be}{0} \hlr{beaten}{1} \hlr{,}{0} \hlr{but}{0} \hlr{it}{0} \hlr{is}{0} \hlr{cheap}{3} \hlr{for}{0} \hlr{a}{0} \hlr{reason}{0} \hlr{.}{0} \hlr{if}{0} \hlr{you}{0} \hlr{come}{0} \hlr{expecting}{0} \hlr{the}{0} \hlr{hilton}{0} \hlr{,}{0} \hlr{then}{0} \hlr{book}{1} \hlr{the}{0} \hlr{hilton}{0} \hlr{!}{0} \hlr{for}{0} \hlr{uk}{0} \hlr{travellers}{0} \hlr{,}{0} \hlr{think}{0} \hlr{of}{0} \hlr{a}{0} \hlr{blackpool}{0} \hlr{b\&b.}{0} \\
		\vspace{0.2mm}
		\emph{Task: Hotel cleanliness}\hspace*{0pt}\hfill R2A-generated attention\\
		\midrule
        \hlr{you}{1} \hlr{get}{1} \hlr{what}{1} \hlr{you}{0} \hlr{pay}{1} \hlr{for}{1} \hlr{.}{0} \hlr{\ul{\textbf{not}}}{0} \hlr{\ul{\textbf{the}}}{1} \hlr{\ul{\textbf{cleanest}}}{20} \hlr{\ul{\textbf{rooms}}}{19} \hlr{\ul{\textbf{but}}}{3} \hlr{\ul{\textbf{bed}}}{7} \hlr{\ul{\textbf{was}}}{1} \hlr{\ul{\textbf{clean}}}{70} \hlr{\ul{\textbf{and}}}{3} \hlr{\ul{\textbf{so}}}{4} \hlr{\ul{\textbf{was}}}{2} \hlr{\ul{\textbf{bathroom}}}{19} \hlr{.}{2} \hlr{bring}{1} \hlr{your}{1} \hlr{own}{1} \hlr{towels}{5} \hlr{though}{1} \hlr{as}{0} \hlr{very}{1} \hlr{thin}{3} \hlr{.}{0} \hlr{service}{0} \hlr{was}{0} \hlr{excellent}{3} \hlr{,}{1} \hlr{let}{0} \hlr{us}{0} \hlr{book}{0} \hlr{in}{0} \hlr{at}{0} \hlr{8:30am}{0} \hlr{!}{0} \hlr{for}{0} \hlr{location}{1} \hlr{and}{0} \hlr{price}{1} \hlr{,}{0} \hlr{this}{0} \hlr{ca}{0} \hlr{n't}{0} \hlr{be}{0} \hlr{beaten}{1} \hlr{,}{0} \hlr{but}{0} \hlr{it}{0} \hlr{is}{0} \hlr{cheap}{3} \hlr{for}{0} \hlr{a}{0} \hlr{reason}{1} \hlr{.}{0} \hlr{if}{0} \hlr{you}{0} \hlr{come}{0} \hlr{expecting}{0} \hlr{the}{0} \hlr{hilton}{0} \hlr{,}{0} \hlr{then}{0} \hlr{book}{0} \hlr{the}{0} \hlr{hilton}{0} \hlr{!}{0} \hlr{for}{0} \hlr{uk}{1} \hlr{travellers}{1} \hlr{,}{0} \hlr{think}{0} \hlr{of}{0} \hlr{a}{0} \hlr{blackpool}{0} \hlr{b\&b.}{0} 
		\\\vspace{0.2mm}
		\emph{Task: Hotel service}\hspace*{0pt}\hfill Oracle Attention\\\midrule
		\hlr{you}{0} \hlr{get}{1} \hlr{what}{1} \hlr{you}{0} \hlr{pay}{8} \hlr{for}{3} \hlr{.}{1} \hlr{not}{0} \hlr{the}{0} \hlr{cleanest}{6} \hlr{rooms}{0} \hlr{but}{0} \hlr{bed}{0} \hlr{was}{0} \hlr{clean}{5} \hlr{and}{0} \hlr{so}{0} \hlr{was}{0} \hlr{bathroom}{1} \hlr{.}{0} \hlr{bring}{3} \hlr{your}{0} \hlr{own}{0} \hlr{towels}{5} \hlr{though}{0} \hlr{as}{0} \hlr{very}{0} \hlr{thin}{5} \hlr{.}{2} \hlr{\ul{\textbf{service}}}{6} \hlr{\ul{\textbf{was}}}{0} \hlr{\ul{\textbf{excellent}}}{70} \hlr{,}{37} \hlr{let}{13} \hlr{us}{4} \hlr{book}{8} \hlr{in}{1} \hlr{at}{0} \hlr{8:30am}{2} \hlr{!}{1} \hlr{for}{0} \hlr{location}{0} \hlr{and}{0} \hlr{price}{2} \hlr{,}{0} \hlr{this}{0} \hlr{ca}{0} \hlr{n't}{0} \hlr{be}{0} \hlr{beaten}{2} \hlr{,}{0} \hlr{but}{0} \hlr{it}{0} \hlr{is}{0} \hlr{cheap}{1} \hlr{for}{0} \hlr{a}{0} \hlr{reason}{1} \hlr{.}{0} \hlr{if}{0} \hlr{you}{0} \hlr{come}{0} \hlr{expecting}{1} \hlr{the}{0} \hlr{hilton}{0} \hlr{,}{0} \hlr{then}{0} \hlr{book}{1} \hlr{the}{0} \hlr{hilton}{0} \hlr{!}{0} \hlr{for}{0} \hlr{uk}{0} \hlr{travellers}{1} \hlr{,}{0} \hlr{think}{0} \hlr{of}{0} \hlr{a}{0} \hlr{blackpool}{1} \hlr{b\&b.}{1} \\
		\vspace{0.2mm}
		\emph{Task: Hotel service}\hspace*{0pt}\hfill R2A-generated attention\\
		\midrule
        \hlr{you}{1} \hlr{get}{2} \hlr{what}{4} \hlr{you}{1} \hlr{pay}{3} \hlr{for}{1} \hlr{.}{1} \hlr{not}{0} \hlr{the}{1} \hlr{cleanest}{1} \hlr{rooms}{2} \hlr{but}{1} \hlr{bed}{1} \hlr{was}{0} \hlr{clean}{5} \hlr{and}{0} \hlr{so}{1} \hlr{was}{1} \hlr{bathroom}{3} \hlr{.}{1} \hlr{bring}{1} \hlr{your}{1} \hlr{own}{1} \hlr{towels}{6} \hlr{though}{2} \hlr{as}{1} \hlr{very}{2} \hlr{thin}{10} \hlr{.}{1} \hlr{\ul{\textbf{service}}}{7} \hlr{\ul{\textbf{was}}}{1} \hlr{\ul{\textbf{excellent}}}{70} \hlr{,}{9} \hlr{let}{5} \hlr{us}{6} \hlr{book}{2} \hlr{in}{1} \hlr{at}{1} \hlr{8:30am}{1} \hlr{!}{1} \hlr{for}{1} \hlr{location}{2} \hlr{and}{1} \hlr{price}{2} \hlr{,}{1} \hlr{this}{0} \hlr{ca}{1} \hlr{n't}{1} \hlr{be}{1} \hlr{beaten}{1} \hlr{,}{1} \hlr{but}{1} \hlr{it}{0} \hlr{is}{0} \hlr{cheap}{6} \hlr{for}{1} \hlr{a}{1} \hlr{reason}{2} \hlr{.}{1} \hlr{if}{1} \hlr{you}{1} \hlr{come}{1} \hlr{expecting}{1} \hlr{the}{1} \hlr{hilton}{1} \hlr{,}{1} \hlr{then}{1} \hlr{book}{0} \hlr{the}{1} \hlr{hilton}{2} \hlr{!}{1} \hlr{for}{1} \hlr{uk}{2} \hlr{travellers}{2} \hlr{,}{1} \hlr{think}{1} \hlr{of}{0} \hlr{a}{0} \hlr{blackpool}{1} \hlr{b\&b.}{1} 
	\end{tabular}
    \captionof{figure}{Oracle attention versus R2A-generated attention on an example of hotel review for three different tasks. Words are highlighted according to the attention scores. Human rationales are shown in bold with underlines. The oracle attention is derived from large amounts of labeled hotel reviews. The R2A is trained on labeled beer reviews with unlabeled hotel reviews.}\label{fig:hotel_res}
\end{table*}

%% file: sections/conclusion.tex
\section{Conclusion}
In this paper, we propose a novel approach that utilizes the connection between human rationales and machine attention to improve the performance of low-resource tasks. Specifically, we learn a transferrable mapping from rationales to high-quality attention on resource-rich tasks. The learned mapping is then used to provide an additional supervision for the target task. Experimental results on both aspect and domain transfer validate that the R2A-generated attention serves as a better form of supervision. Our model produces high-quality attention for low-resource tasks.

%% file: sections/appendix.tex
\clearpage

\section*{A1 Generating machine rationales}\label{sec:a1}
Collecting rationales at scale is expensive. Thus, we employ an existing rationalization model~\cite{lei2016rationalizing} to generate high-quality rationales automatically for the resource-rich source tasks. 

The rationalization model is composed of two modular components: a generator and a classifier. The generator generates rationales from the input, and the classifier maps the generated rationales into the final label. The two components are trained jointly to minimize a loss function that favors short, concise rationales while enforcing that the rationales alone suffice for accurate prediction.

\begin{figure}[b!]
    \centering
    \includegraphics[width=.8\linewidth]{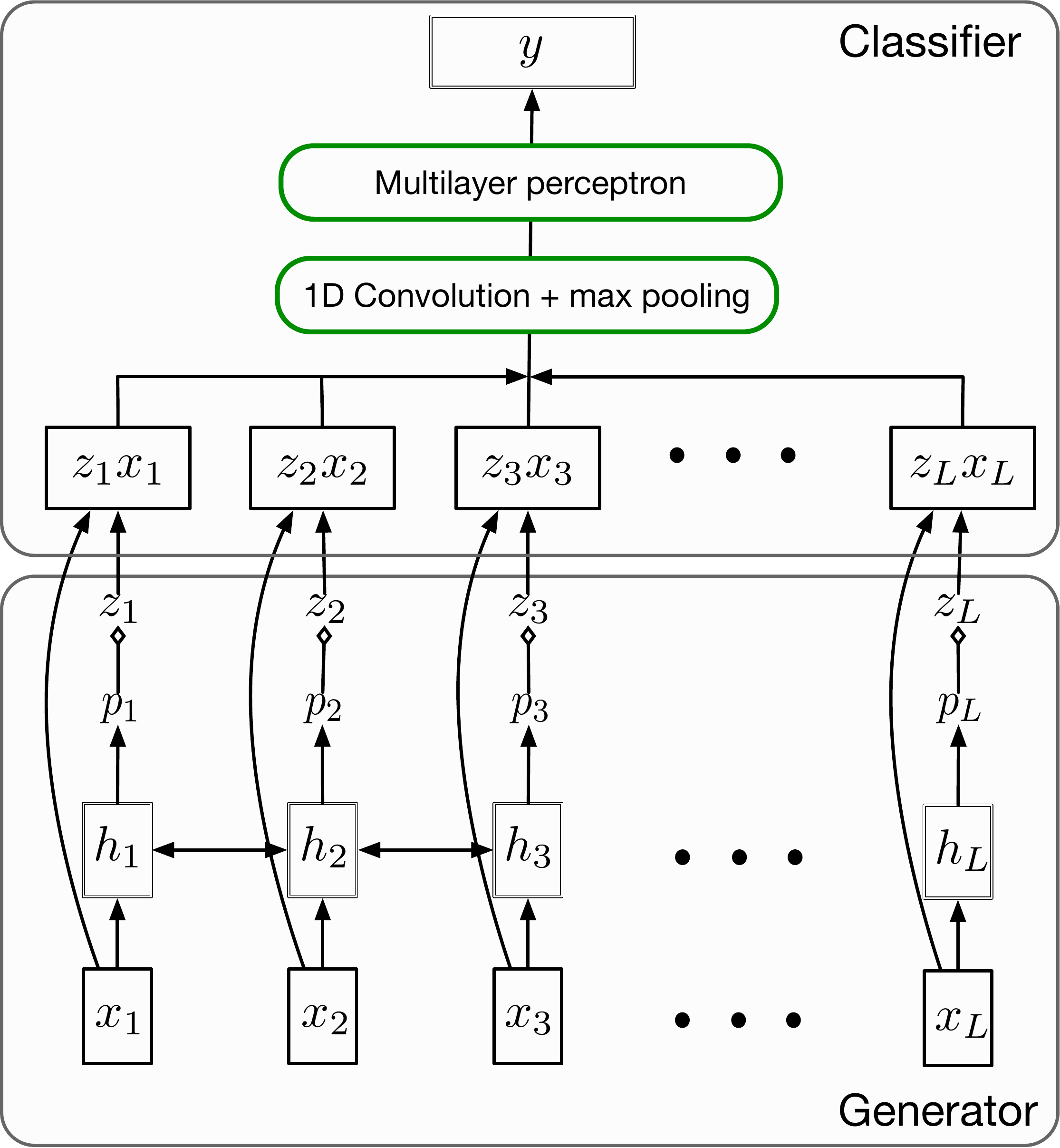}
    \caption{Rationalization model~\cite{lei2016rationalizing}. Arrow line denotes deterministic computations, while diamond line denotes stochastic sampling.}
    \label{fig:tao}
\end{figure}

Figure~\ref{fig:tao} illustrates the model architecture. For the generator, we use a 200 dimensional bi-LSTM to encode the word embedding sequence. Then we apply a linear regressor at each position $i$ to predict the probability $p_i$ that the current word is a rationale. We sample $z_i$ from this probability and pass the sampled rationales $z_ix_i$ to the classifier module.  To encourage fast and stable convergence, we use the Gumbel trick~\cite{jang2016categorical} during sampling.

For the classifier, we employ a CNN-based model~\cite{kim2014convolutional}. Specifically, the classifier first computes 1D convolution over the embeddings of the generated rationales. The filter windows are 3, 5, 7 with 50 feature maps each. It then applies max-over-time pooling to obtain a fixed-length feature vector. Finally, a multilayer perceptron (MLP) is used to predict the label from the feature vector. The MLP consists of one hidden layer (50 units and ReLU activation).

\section*{A2 Collecting human rationales}\label{sec:a2}
\begin{figure}[t]
\centering
\includegraphics[width=\linewidth]{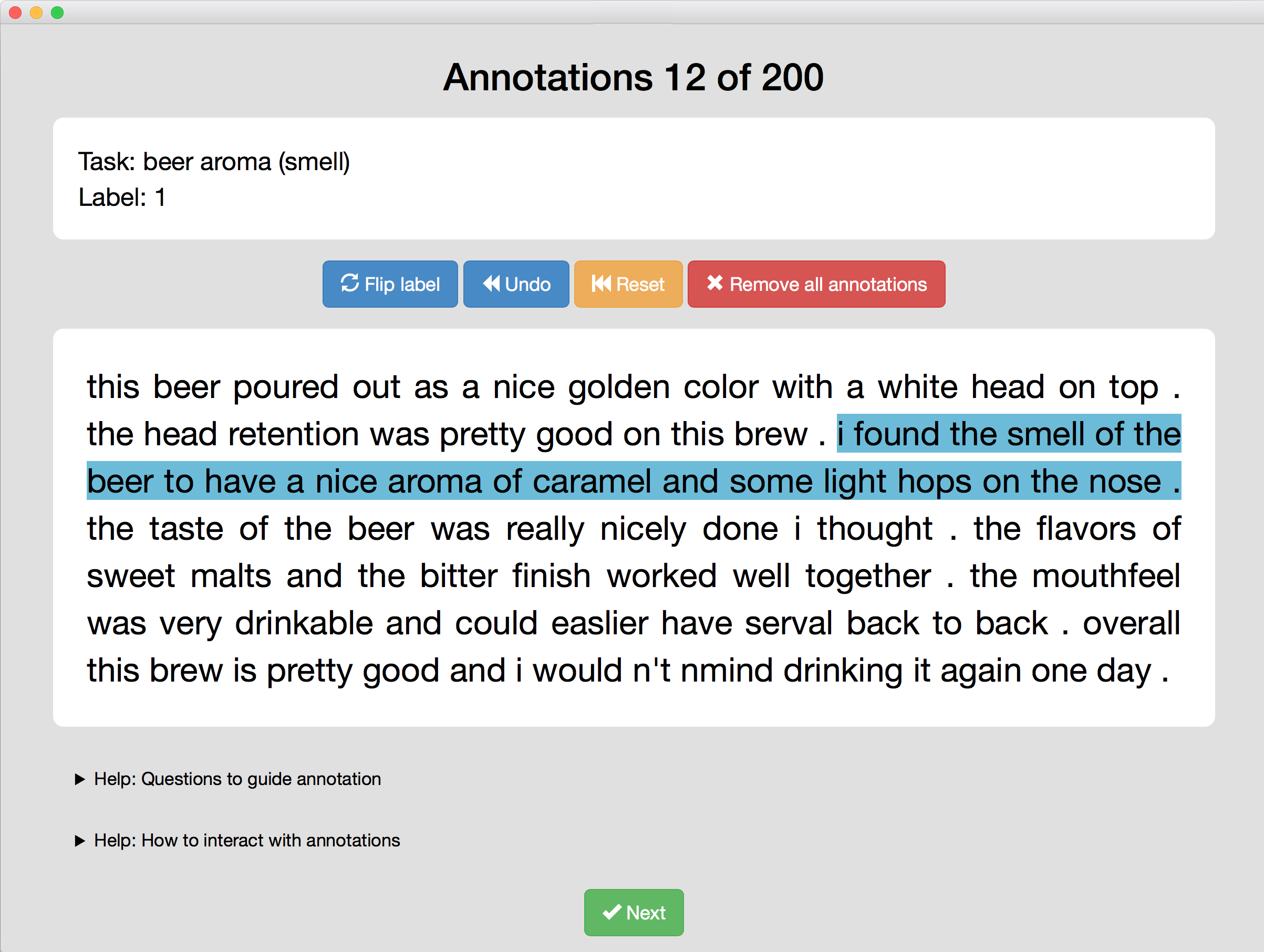}
\caption{Screenshot of the annotation tool}\label{fig:annotation}
\end{figure}

We collected human rationales on six sentiment classification tasks: beer look, beer aroma, beer palate, hotel location, hotel cleanliness and hotel service. For each task, we randomly picked 100 positive examples and 100 negative examples. These 200 labeled examples are given to the annotators (five students) to highlight rationales that are short and coherent, yet sufficient for supporting the label~\cite{lei2016rationalizing}. The annotators can also flip the original label if it is incorrect. Figure~\ref{fig:annotation} shows our annotation interface, and Table~\ref{tab:rationale_stats} presents the statistics of the collected rationales.

\begin{table}[t]
\centering
\scriptsize
\begin{tabular}{lcccc}
\toprule
Task &  \shortstack[c]{\#tokens\\ per review} & \shortstack[c]{\#rationales\\ per review} & \shortstack[c]{\#unique\\tokens}& \shortstack[c]{\#unique\\ rationales} \\
\midrule
Beer look & 125.4 & 21.4 & 35,025 & 6,460 \\
\midrule
Beer aroma & 126.3 & 17.6 & 35,088 & 5,611 \\
\midrule
Beer palate & 133.5 & 13.0 & 36,786 & 3,885  \\
\midrule
Hotel location & 157.1 & 13.4 & 42,717 & 4,174 \\
\midrule
Hotel cleanliness & 149.4 & 13.2 & 42,045 & 4,758 \\
\midrule
Hotel service & 154.5 & 17.5 & 42,989 & 6,110 \\
\bottomrule
\end{tabular}
\caption{Statistics of the annotated rationales.}\label{tab:rationale_stats}
\end{table}

\begin{table*}[t!]
\centering\small
\begin{tabular}{lccccccc}
\toprule
     &\multicolumn{3}{c}{Beer} & \multicolumn{3}{c}{Hotel}  \\
      \cmidrule(lr){2-4}  \cmidrule(lr){5-7}
      & Look & Aroma & Palate & Location & Cleanliness & Service\\
      \midrule
     \#training data & 32,276 & 28,984 & 25,748 & 14,472 & 150,098 & 101,484 \\
     \#testing data & 4,014 & 4,212 & 3,804 & 1,808 & 12,684 & 18,762 \\
     \midrule
     Testing accuracy & 87.17 & 86.35 & 82.02 & 92.20 & 94.62 & 95.17 \\
\bottomrule
\end{tabular}
\caption{Data used to derive the oracle attention. The accuracy is evaluated on the testing data shown in Table~\ref{tab:data_beer} and \ref{tab:data_hotel}. }\label{tab:oracledata}
\end{table*}

\section*{A3 Deriving the oracle attention}\label{sec:a3}
\begin{figure}[t]
    \centering
    \includegraphics[width=.8\linewidth]{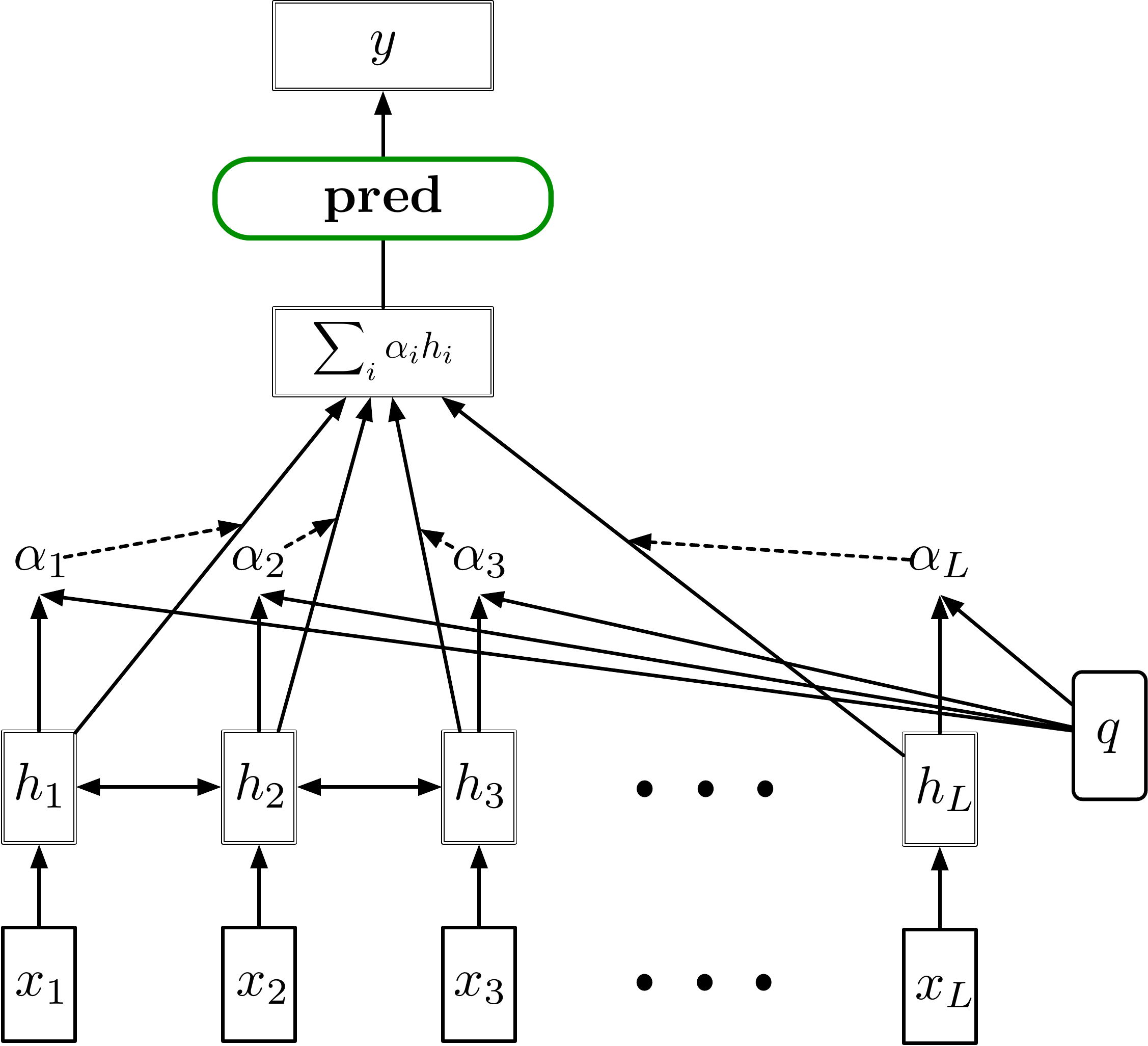}
    \caption{Attention-based classifier}
    \label{fig:attclf}
\end{figure}

Figure~\ref{fig:attclf} illustrates the model architecture that we used to derive the oracle attention. It has the same architecture as the source classifier in our \rtoa\ model (Section~\ref{ssec:meta_learn}). Specifically, we use pre-trained fastText embeddings. The encoder is a bi-LSTM with 200 hidden units. The attention head $q$ is a vector of dimension $50$. For the prediction module $\mathrm{\mathbf{pred}}$, we use a MLP with one hidden layer (50 units and ReLU activation). 
For each task, we train this attention-based classifier on a large amount of annotations. Table~\ref{tab:oracledata} presents the statistics of the training data and the testing performance of the classifier.

\section*{A4 Estimating the Wasserstein distance}\label{sec:a4}

\begin{figure}[t!]
    \centering
    \includegraphics[width=\linewidth]{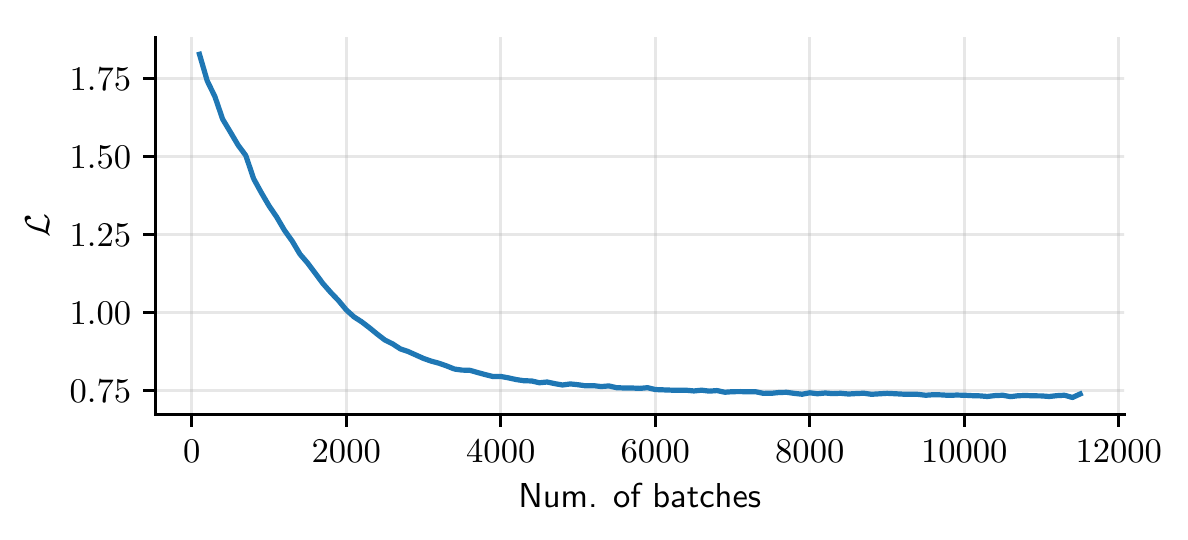}
    \caption{Domain transfer from beer review to hotel review. Overall loss $\mathcal{L}$ of \rtoa\ on the source development set during training.}
    \label{fig:iter}
\end{figure}

We train a critic network to estimate the Wasserstein distance between the  distribution of the representation from the source domain and the one from the target domain:
\begin{equation*}
    \begin{aligned}
    \mathcal{L}_{wd} = \sup_{\|f\|_{L}\leq K}
    &\mathbb{E}_{h^{\rmb{inv}} \sim
    \mathbb{P}_{\mathcal{S}}}\left[f([h^{\rmb{inv}}_1; h^{\rmb{inv}}_{L}])\right]\\
    \quad &-\mathbb{E}_{h^{\rmb{inv}} \sim
    \mathbb{P}_{\mathcal{T}}}\left[f([h^{\rmb{inv}}_1; h^{\rmb{inv}}_{L}])\right].
    \end{aligned}
\end{equation*}
The critic network is parametrized as a MLP with one hidden layer (100 units  and ReLU activation). Following \citet{gulrajani2017improved}, we set the weight of the gradient penalty to 10 and optimize the critic network for 5 iterations during each batch. Figure~\ref{fig:iter} plots the loss on the development set versus the number of training batches. We see that our \rtoa\ model converges smoothly during training.

\quad\\
\quad\\
\quad\\
\quad\\
\quad\\

%% file: r2a.bbl
\begin{thebibliography}{34}
\expandafter\ifx\csname natexlab\endcsname\relax\def\natexlab#1{#1}\fi

\bibitem[{Arjovsky et~al.(2017)Arjovsky, Chintala, and
  Bottou}]{arjovsky2017wasserstein}
Martin Arjovsky, Soumith Chintala, and L{\'e}on Bottou. 2017.
\newblock Wasserstein gan.
\newblock \emph{arXiv preprint arXiv:1701.07875}.

\bibitem[{Bahdanau et~al.(2014)Bahdanau, Cho, and Bengio}]{bahdanau2014neural}
Dzmitry Bahdanau, Kyunghyun Cho, and Yoshua Bengio. 2014.
\newblock Neural machine translation by jointly learning to align and
  translate.
\newblock \emph{arXiv preprint arXiv:1409.0473}.

\bibitem[{Bengio et~al.(2003)Bengio, Ducharme, Vincent, and
  Jauvin}]{bengio2003neural}
Yoshua Bengio, R{\'e}jean Ducharme, Pascal Vincent, and Christian Jauvin. 2003.
\newblock A neural probabilistic language model.
\newblock \emph{Journal of machine learning research}, 3(Feb):1137--1155.

\bibitem[{Bojanowski et~al.(2017)Bojanowski, Grave, Joulin, and
  Mikolov}]{bojanowski2017enriching}
Piotr Bojanowski, Edouard Grave, Armand Joulin, and Tomas Mikolov. 2017.
\newblock Enriching word vectors with subword information.
\newblock \emph{Transactions of the Association for Computational Linguistics},
  5:135--146.

\bibitem[{Chen et~al.(2017)Chen, Fisch, Weston, and Bordes}]{chen2017reading}
Danqi Chen, Adam Fisch, Jason Weston, and Antoine Bordes. 2017.
\newblock Reading wikipedia to answer open-domain questions.
\newblock In \emph{Proceedings of the 55th Annual Meeting of the Association
  for Computational Linguistics (Volume 1: Long Papers)}, volume~1, pages
  1870--1879.

\bibitem[{Chen et~al.(2012)Chen, Xu, Weinberger, and
  Sha}]{chen2012marginalized}
Minmin Chen, Zhixiang Xu, Kilian Weinberger, and Fei Sha. 2012.
\newblock Marginalized denoising autoencoders for domain adaptation.
\newblock \emph{arXiv preprint arXiv:1206.4683}.

\bibitem[{Collobert et~al.(2011)Collobert, Weston, Bottou, Karlen, Kavukcuoglu,
  and Kuksa}]{collobert2011natural}
Ronan Collobert, Jason Weston, L{\'e}on Bottou, Michael Karlen, Koray
  Kavukcuoglu, and Pavel Kuksa. 2011.
\newblock Natural language processing (almost) from scratch.
\newblock \emph{Journal of Machine Learning Research}, 12(Aug):2493--2537.

\bibitem[{Conneau et~al.(2017)Conneau, Kiela, Schwenk, Barrault, and
  Bordes}]{conneau2017supervised}
Alexis Conneau, Douwe Kiela, Holger Schwenk, Loic Barrault, and Antoine Bordes.
  2017.
\newblock Supervised learning of universal sentence representations from
  natural language inference data.
\newblock \emph{arXiv preprint arXiv:1705.02364}.

\bibitem[{Ganin et~al.(2016)Ganin, Ustinova, Ajakan, Germain, Larochelle,
  Laviolette, Marchand, and Lempitsky}]{ganin2016domain}
Yaroslav Ganin, Evgeniya Ustinova, Hana Ajakan, Pascal Germain, Hugo
  Larochelle, Fran{\c{c}}ois Laviolette, Mario Marchand, and Victor Lempitsky.
  2016.
\newblock Domain-adversarial training of neural networks.
\newblock \emph{The Journal of Machine Learning Research}, 17(1):2096--2030.

\bibitem[{Glorot et~al.(2011)Glorot, Bordes, and Bengio}]{glorot2011domain}
Xavier Glorot, Antoine Bordes, and Yoshua Bengio. 2011.
\newblock Domain adaptation for large-scale sentiment classification: A deep
  learning approach.
\newblock In \emph{Proceedings of the 28th international conference on machine
  learning (ICML-11)}, pages 513--520.

\bibitem[{Gulrajani et~al.(2017)Gulrajani, Ahmed, Arjovsky, Dumoulin, and
  Courville}]{gulrajani2017improved}
Ishaan Gulrajani, Faruk Ahmed, Martin Arjovsky, Vincent Dumoulin, and Aaron~C
  Courville. 2017.
\newblock Improved training of wasserstein gans.
\newblock In \emph{Advances in Neural Information Processing Systems}, pages
  5769--5779.

\bibitem[{Hochreiter and Schmidhuber(1997)}]{hochreiter1997long}
Sepp Hochreiter and J{\"u}rgen Schmidhuber. 1997.
\newblock Long short-term memory.
\newblock \emph{Neural computation}, 9(8):1735--1780.

\bibitem[{Jang et~al.(2016)Jang, Gu, and Poole}]{jang2016categorical}
Eric Jang, Shixiang Gu, and Ben Poole. 2016.
\newblock Categorical reparameterization with gumbel-softmax.
\newblock \emph{arXiv preprint arXiv:1611.01144}.

\bibitem[{Kamigaito et~al.(2017)Kamigaito, Hayashi, Hirao, Takamura, Okumura,
  and Nagata}]{kamigaito2017supervised}
Hidetaka Kamigaito, Katsuhiko Hayashi, Tsutomu Hirao, Hiroya Takamura, Manabu
  Okumura, and Masaaki Nagata. 2017.
\newblock Supervised attention for sequence-to-sequence constituency parsing.
\newblock In \emph{Proceedings of the Eighth International Joint Conference on
  Natural Language Processing (Volume 2: Short Papers)}, volume~2, pages 7--12.

\bibitem[{Kim(2014)}]{kim2014convolutional}
Yoon Kim. 2014.
\newblock Convolutional neural networks for sentence classification.
\newblock \emph{arXiv preprint arXiv:1408.5882}.

\bibitem[{Kingma and Ba(2014)}]{kingma2014adam}
Diederik~P Kingma and Jimmy Ba. 2014.
\newblock Adam: A method for stochastic optimization.
\newblock \emph{arXiv preprint arXiv:1412.6980}.

\bibitem[{Lei et~al.(2016)Lei, Barzilay, and Jaakkola}]{lei2016rationalizing}
Tao Lei, Regina Barzilay, and Tommi Jaakkola. 2016.
\newblock Rationalizing neural predictions.
\newblock In \emph{Proceedings of the 2016 Conference on Empirical Methods in
  Natural Language Processing}, pages 107--117.

\bibitem[{Lin et~al.(2017)Lin, Feng, Santos, Yu, Xiang, Zhou, and
  Bengio}]{lin2017structured}
Zhouhan Lin, Minwei Feng, Cicero Nogueira~dos Santos, Mo~Yu, Bing Xiang, Bowen
  Zhou, and Yoshua Bengio. 2017.
\newblock A structured self-attentive sentence embedding.
\newblock \emph{arXiv preprint arXiv:1703.03130}.

\bibitem[{Liu et~al.(2016)Liu, Utiyama, Finch, and Sumita}]{liu2016neural}
Lemao Liu, Masao Utiyama, Andrew Finch, and Eiichiro Sumita. 2016.
\newblock Neural machine translation with supervised attention.
\newblock \emph{arXiv preprint arXiv:1609.04186}.

\bibitem[{Liu et~al.(2017)Liu, Chen, Liu, and Zhao}]{liu2017exploiting}
Shulin Liu, Yubo Chen, Kang Liu, and Jun Zhao. 2017.
\newblock Exploiting argument information to improve event detection via
  supervised attention mechanisms.
\newblock In \emph{Proceedings of the 55th Annual Meeting of the Association
  for Computational Linguistics (Volume 1: Long Papers)}, volume~1, pages
  1789--1798.

\bibitem[{Luong et~al.(2015)Luong, Pham, and Manning}]{luong2015effective}
Thang Luong, Hieu Pham, and Christopher~D Manning. 2015.
\newblock Effective approaches to attention-based neural machine translation.
\newblock In \emph{Proceedings of the 2015 Conference on Empirical Methods in
  Natural Language Processing}, pages 1412--1421.

\bibitem[{Maaten and Hinton(2008)}]{maaten2008visualizing}
Laurens van~der Maaten and Geoffrey Hinton. 2008.
\newblock Visualizing data using t-sne.
\newblock \emph{Journal of machine learning research}, 9(Nov):2579--2605.

\bibitem[{McAuley et~al.(2012)McAuley, Leskovec, and
  Jurafsky}]{mcauley2012learning}
Julian McAuley, Jure Leskovec, and Dan Jurafsky. 2012.
\newblock Learning attitudes and attributes from multi-aspect reviews.
\newblock In \emph{Data Mining (ICDM), 2012 IEEE 12th International Conference
  on}, pages 1020--1025. IEEE.

\bibitem[{Mikolov et~al.(2011)Mikolov, Kombrink, Burget, {\v{C}}ernock{\`y},
  and Khudanpur}]{mikolov2011extensions}
Tom{\'a}{\v{s}} Mikolov, Stefan Kombrink, Luk{\'a}{\v{s}} Burget, Jan
  {\v{C}}ernock{\`y}, and Sanjeev Khudanpur. 2011.
\newblock Extensions of recurrent neural network language model.
\newblock In \emph{Acoustics, Speech and Signal Processing (ICASSP), 2011 IEEE
  International Conference on}, pages 5528--5531. IEEE.

\bibitem[{Peters et~al.(2018)Peters, Neumann, Iyyer, Gardner, Clark, Lee, and
  Zettlemoyer}]{peters2018deep}
Matthew~E Peters, Mark Neumann, Mohit Iyyer, Matt Gardner, Christopher Clark,
  Kenton Lee, and Luke Zettlemoyer. 2018.
\newblock Deep contextualized word representations.
\newblock \emph{arXiv preprint arXiv:1802.05365}.

\bibitem[{Rush et~al.(2015)Rush, Chopra, and Weston}]{rush2015neural}
Alexander~M Rush, Sumit Chopra, and Jason Weston. 2015.
\newblock A neural attention model for abstractive sentence summarization.
\newblock \emph{arXiv preprint arXiv:1509.00685}.

\bibitem[{Srivastava et~al.(2014)Srivastava, Hinton, Krizhevsky, Sutskever, and
  Salakhutdinov}]{srivastava2014dropout}
Nitish Srivastava, Geoffrey Hinton, Alex Krizhevsky, Ilya Sutskever, and Ruslan
  Salakhutdinov. 2014.
\newblock Dropout: A simple way to prevent neural networks from overfitting.
\newblock \emph{The Journal of Machine Learning Research}, 15(1):1929--1958.

\bibitem[{Vaswani et~al.(2017)Vaswani, Shazeer, Parmar, Uszkoreit, Jones,
  Gomez, Kaiser, and Polosukhin}]{NIPS2017_7181}
Ashish Vaswani, Noam Shazeer, Niki Parmar, Jakob Uszkoreit, Llion Jones,
  Aidan~N Gomez, \L~ukasz Kaiser, and Illia Polosukhin. 2017.
\newblock Attention is all you need.
\newblock In I.~Guyon, U.~V. Luxburg, S.~Bengio, H.~Wallach, R.~Fergus,
  S.~Vishwanathan, and R.~Garnett, editors, \emph{Advances in Neural
  Information Processing Systems 30}, pages 5998--6008. Curran Associates, Inc.

\bibitem[{Wang et~al.(2010)Wang, Lu, and Zhai}]{wang2010latent}
Hongning Wang, Yue Lu, and Chengxiang Zhai. 2010.
\newblock Latent aspect rating analysis on review text data: a rating
  regression approach.
\newblock In \emph{Proceedings of the 16th ACM SIGKDD international conference
  on Knowledge discovery and data mining}, pages 783--792. ACm.

\bibitem[{Yang et~al.(2016)Yang, Yang, Dyer, He, Smola, and
  Hovy}]{yang2016hierarchical}
Zichao Yang, Diyi Yang, Chris Dyer, Xiaodong He, Alex Smola, and Eduard Hovy.
  2016.
\newblock Hierarchical attention networks for document classification.
\newblock In \emph{Proceedings of the 2016 Conference of the North American
  Chapter of the Association for Computational Linguistics: Human Language
  Technologies}, pages 1480--1489.

\bibitem[{Zaidan et~al.(2007)Zaidan, Eisner, and Piatko}]{zaidan2007using}
Omar Zaidan, Jason Eisner, and Christine Piatko. 2007.
\newblock Using “annotator rationales” to improve machine learning for text
  categorization.
\newblock In \emph{Human Language Technologies 2007: The Conference of the
  North American Chapter of the Association for Computational Linguistics;
  Proceedings of the Main Conference}, pages 260--267.

\bibitem[{Zhang et~al.(2016)Zhang, Marshall, and Wallace}]{zhang2016rationale}
Ye~Zhang, Iain Marshall, and Byron~C Wallace. 2016.
\newblock Rationale-augmented convolutional neural networks for text
  classification.
\newblock In \emph{Proceedings of the Conference on Empirical Methods in
  Natural Language Processing. Conference on Empirical Methods in Natural
  Language Processing}, volume 2016, page 795. NIH Public Access.

\bibitem[{Zhang et~al.(2017)Zhang, Barzilay, and Jaakkola}]{zhang2017aspect}
Yuan Zhang, Regina Barzilay, and Tommi Jaakkola. 2017.
\newblock Aspect-augmented adversarial networks for domain adaptation.
\newblock \emph{Transactions of the Association of Computational Linguistics},
  5(1):515--528.

\bibitem[{Zhou et~al.(2016)Zhou, Xie, Huang, and He}]{zhou2016bi}
Guangyou Zhou, Zhiwen Xie, Jimmy~Xiangji Huang, and Tingting He. 2016.
\newblock Bi-transferring deep neural networks for domain adaptation.
\newblock In \emph{Proceedings of the 54th Annual Meeting of the Association
  for Computational Linguistics (Volume 1: Long Papers)}, volume~1, pages
  322--332.

\end{thebibliography}
